\documentclass[nonacm,sigplan]{acmart}

\usepackage[]{hyperref}

\usepackage[normalem]{ulem}

\usepackage{multirow}
\usepackage[ruled,vlined,linesnumbered]{algorithm2e}
\usepackage{float}
\usepackage{mathtools}
\usepackage{comment}
\usepackage{enumitem}
\usepackage{amsthm}
\usepackage{subfig}
\usepackage{makecell}
\usepackage{booktabs}

\setlist[itemize]{leftmargin=15pt,itemsep=0em,topsep=0em}

\DeclareMathAlphabet{\mathcal}{OMS}{cmsy}{m}{n}

\sfcode`\.=1001
\sfcode`\?=1001
\sfcode`\!=1001
\sfcode`\:=1001

\newcommand\hmm[1]{\ifnum\spacefactor=1001 \uppercase{#1}\else#1\fi}
\newcommand{\MixGCN}{\textsc{MixGCN}\xspace}
\newcommand{\MoP}{\textit{MoP}\xspace}
\newcommand{\MoA}{\textit{MoA}\xspace}

\newcommand{\fullMoA}{\textit{\hmm{m}ixture of accelerators}\xspace}

\newcommand{\niparagraph}[1]{\noindent\textbf{#1.}}

\newtheorem{proposition}{Proposition}[section]

\SetKwComment{Comment}{$\triangleright$\ }{}

\SetCommentSty{mycommfont}

\graphicspath{{figure/}}

\begin{document}

\title{\MixGCN: Scalable GCN Training by Mixture of Parallelism and Mixture of Accelerators}

\author{Cheng Wan}
\email{chwan@gatech.edu}
\affiliation{%
  \institution{Georgia Institute of Technology}
  \city{Atlanta}
  \state{Georgia}
  \country{USA}
}

\author{Runkai Tao}
\email{rt572@rutgers.edu}
\affiliation{%
  \institution{Rutgers University}
  \city{New Brunswick}
  \state{New Jersey}
  \country{USA}}

\author{Zheng Du}
\email{zhaolab@umn.edu}
\affiliation{%
 \institution{University of Minnesota Twin Cities}
 \city{Minneapolis}
 \state{Minnesota}
 \country{USA}
}

\author{Yang (Katie) Zhao}
\email{yangzhao@umn.edu}
\affiliation{%
 \institution{University of Minnesota Twin Cities}
 \city{Minneapolis}
 \state{Minnesota}
 \country{USA}
}

\author{Yingyan (Celine) Lin}
\email{celine.lin@gatech.edu }
\affiliation{%
  \institution{Georgia Institute of Technology}
  \city{Atlanta}
  \state{Georgia}
  \country{USA}
}

\begin{abstract}
Graph convolutional networks (GCNs) have demonstrated superiority in graph-based learning tasks. However, training GCNs on full graphs is particularly challenging, due to the following two challenges: (1) the associated feature tensors can easily explode the memory and block the communication bandwidth of modern accelerators, and (2) the computation workflow in training GCNs alternates between sparse and dense matrix operations, complicating the efficient utilization of computational resources. Existing solutions for scalable distributed full-graph GCN training mostly adopt partition parallelism, which is unsatisfactory as they only partially address the first challenge while incurring scaled-out communication volume. To this end, we propose \MixGCN aiming to simultaneously address both the aforementioned challenges towards GCN training. To tackle the first challenge, \MixGCN integrates \textit{mixture of parallelism}. Both theoretical and empirical analysis verify its constant communication volumes and enhanced balanced workload; For handling the second challenge, we consider \textit{mixture of accelerators} (i.e., sparse and dense accelerators) with a dedicated accelerator for GCN training and a fine-grain pipeline. Extensive experiments show that \MixGCN achieves boosted training efficiency and scalability. 
\end{abstract}

\maketitle 
\pagestyle{plain} 

\section{Introduction}

Graphs have served as a natural representation of real-world data thanks to its ability of depicting dependent relationship. 
Learning over graphs has been a popular research topic for the past decades \cite{jiang2013survey,chen2020graph,wu2020comprehensive,xia2021graph,ji2021survey}, and one recent emerging method is graph convolutional networks (GCNs) \cite{kipf2016semi}, which enjoys powerful expressive capabilities \cite{xu2018powerful,wijesinghe2021new} and has been adopted to various real-world applications \cite{huang2020recurrent,ying2018graph,guo2022dataefficient,wan2024towards}. 
Specifically, the computation of a GCN follows a two-step process: \textit{neighbor aggregation} and \textit{node update}. For a given node, to calculate its features in the next layer, a GCN first leverages a permutation-invariant function (e.g., average pooling) for aggregating all features from its incoming neighbor set, and then utilizes an update function (e.g., a multilayer perceptron) to combine the aggregated features and the embedding vector in the previous layer to calculate the new vector representation of the target node. 
Such a two-step process allows GCNs to capture the structure of input graphs and further retain the powerful expressive capabilities of neural networks.

\begin{figure*}[t]
\centering
    \subfloat[An input graph.]{
        \includegraphics[width=0.21\linewidth]{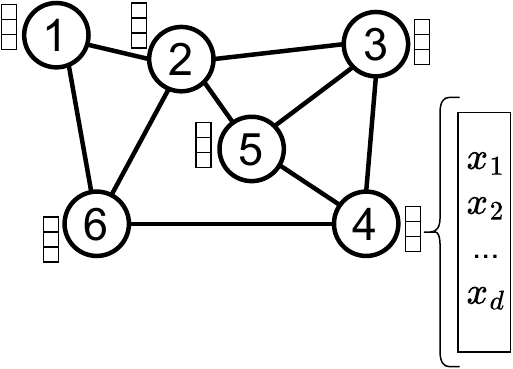}
        \label{fig:gcn_comparison_a}
    }
    \subfloat[Overview of partition parallelism.]{
        \includegraphics[width=0.31\linewidth]{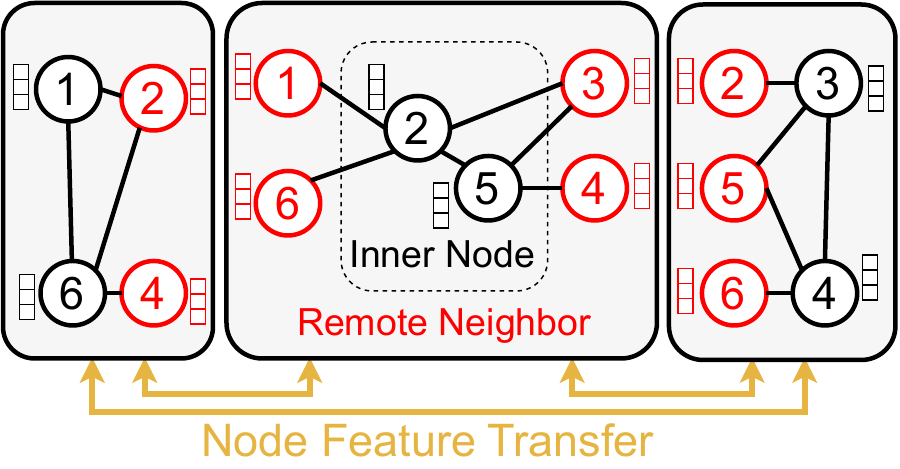}
        \label{fig:gcn_comparison_b}
    }
    \subfloat[Overview of the proposed \MixGCN.]{
        \includegraphics[width=0.44\linewidth]{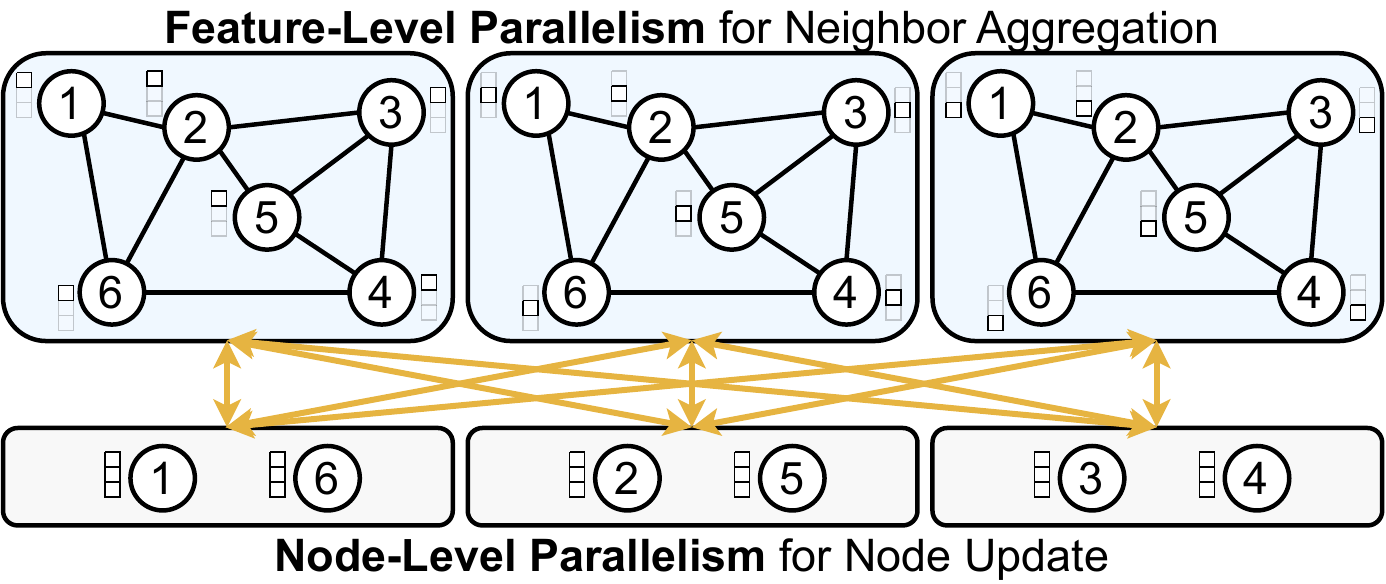}
        \label{fig:gcn_comparison_c}
    }
\caption{An illustrative comparison between partition parallelism and the proposed \MixGCN, where \MixGCN 
avoids the scaled-out communication volume needed for duplicated remote neighbor features (highlighted in red in (b)) as required by partition parallelism.}
\label{fig:gcn_comparison}
\end{figure*}

Despite GCNs' great potential, designing scalable GCN training systems is particularly challenging and still under-explored, due to the associated \textit{giant feature tensors}. 
For example, ogbn-papers100M~\cite{hu2020open}, a popular dataset for GCN research, contains more than 100 million nodes, requiring 124GB for storing merely the features and labels, let alone the storage requirement for storing the intermediate features to support backward propagation, which cannot be fit in a single modern accelerator.
As such, to handle large-graph training, many recent works follow the direction of partition parallelism \cite{ma2019neugraph,zhu2019aligraph,jia2020improving,zheng2020distdgl,wang2021flexgraph,tripathy2020reducing,md2021distgnn,fey2021gnnautoscale,thorpe2021dorylus,gandhi2021p3,wang2020gnnadvisor,wan2022bns,wan2022pipegcn,ramezani2021learn,peng2022sancus,wang2022neutronstar,chai2022distributed,bai2023staleness,zhang2024sylvie,huang2024wisegraph,wan2023adaptive}, as depicted in Figure~\ref{fig:gcn_comparison_b}. 
The key idea is to separate a giant graph into multiple partitions, and assign each partition to one single accelerator. This straightforward approach, however, only distributes the storage of feature tensors, while incurring a significant memory and communication overhead for duplicating the remote neighbors from other accelerators (i.e., the red nodes in Figure~\ref{fig:gcn_comparison_b}) \cite{wan2022bns}. This has restricted the scalability of GCN training due to the scaled-out number of remote neighbors. Furthermore, as we will show in Section~\ref{sec:mop_pp}, balancing workload via partition parallelism is NP-Hard, leading to a nontrivial synchronization overhead.

In parallel, the existing systems for scalable deep neural networks (DNNs) training can not be adopted for scalable GCN training. 
This is because these systems do not consider and thus are not optimized for handling the unique GCN training workflow which consists of \textit{hybrid sparse-dense operations}. 
In particular, training GCNs alternatively performs sparse matrix operations for \textit{neighbor aggregation} and dense matrix operations for \textit{node update}. 
While the involved sparse-dense operations strengthen the capability of GCNs, they do not suit the underlying design of modern distributed systems for DNN training, of which the workflow is composed of dependent dense operations \cite{chen2015mxnet,abadi2016tensorflow,li2020pytorch}. Consequently, a straightforward deployment of GCN training into a distributed DNN system would suffer from low hardware utilization and inefficiency. 

In summary, there exist two unique challenges associated with GCN training on large-scale graphs: \textit{giant feature tensors} and
\textit{hybrid sparse-dense operations}, severely 
challenging the design of efficient and scalable GCN training systems. 
We propose \MixGCN to handle these challenges.

\textbf{Contribution 1: On the system level, we propose \textit{Mixture of Parallelism} (\MoP).} For addressing the training inefficiency caused by \textit{giant feature tensors}, we develop \textit{Mixture of Parallelism} (\MoP), a hybrid feature- and node-level parallelism to improve the scalability of training GCNs. As demonstrated in Figure~\ref{fig:gcn_comparison_c}, 
\MixGCN leverages feature-level parallelism for \textit{neighbor aggregation} and node-level parallelism for \textit{node update}. 
Compared with partition parallelism that induces scaled-out communication volume, \MoP avoids the necessity of duplicating remote neighbors and thus requires constant communication volumes. 
We describe our \MoP technique and detailed analysis in Section~\ref{sec:mop}.

\textbf{Contribution 2: On the architecture level, we propose \textit{Mixture of Accelerators} (\MoA).} 
Thanks to our \MoP technique, the second unique challenge associated with GCN training, i.e., \textit{hybrid sparse-dense operations} can be naturally assigned to different sets of accelerators. We thus design a novel distributed training system, \textit{Mixture of Accelerators} (\MoA), where the sparse accelerators (i.e., the blue parts in Figure~\ref{fig:gcn_comparison_c}) tackle sparse matrix operations (i.e., \textit{neighbor aggregation}) while the dense accelerators (i.e., the grey parts in Figure~\ref{fig:gcn_comparison_c}) are responsible for handling 
the dense matrix operations (i.e., \textit{node update}). 
We further identify a unique sparse operation named S-SpMM in GNN training, which fuses two consecutive sparse operations, and devise a dedicated accelerator for efficient computation. 
A fine-grain pipeline with node reordering is adopted to further enhance scalability.
Section~\ref{sec:moa} provides more details. 

\textbf{Contribution 3: By combining the two proposed techniques above, we validate the performance of \MixGCN}. Extensive experiments over 5 large-scale datasets verify that \MixGCN offers multiple advantages simultaneously, as detailed in Section \ref{sec:experiment}.
\begin{itemize}
    \item End-to-end empirical evaluations demonstrate that \MixGCN enjoys the highest end-to-end throughput, outperforming state-of-the-art baselines by 10.4$\times$ on a 4-node GPU cluster. Simulation results indicate that this performance gain can be further increased to 17.2$\times$ when utilizing our dedicated sparse accelerator (Section~\ref{sec:exp_e2e}).
    \item Profiling results confirm that the proposed \MoP maintains a constant communication volume and feature memory usage, while ensuring a fully balanced workload (Section~\ref{sec:exp_mop}).
    \item Detailed ablation studies show that our dedicated accelerator surpasses existing GCN accelerators, delivering up to a 3.4$\times$ speedup. In parallel, a fine-grain pipeline with node reordering leads to a speedup of 1.18$\times$. (Section~\ref{sec:exp_moa}).
\end{itemize}
\section{Background and Related Work}
\subsection{Graph Convolutional Networks}
\label{gcn_background}
GCNs are popular for graph-based learning tasks. Each layer of a GCN uses a two-step process to calculate the new feature embedding of each node, which can be represented as:
\begin{align}
z^{(l)}_v&=\zeta^{(l)}\left(\left\{h^{(l-1)}_u\mid u\in\mathcal{N}(v)\right\}\right) \label{eq:aggr} \\
h^{(l)}_v&=\phi^{(l)}\left(z^{(l)}_v,h^{(l-1)}_v\right)\label{eq:update}
\end{align}
where $\mathcal{N}(v)$ represents the neighbor set of node $v$, $h^{(l)}_v$ is the feature vector of node $v$ calculated by the $l$-th layer, $\zeta^{(l)}$ denotes an aggregation function for calculating the intermediate result $z_v^{(l)}$, and $\phi^{(l)}$ denotes an update function for updating the features of each node. We call the process of Equation~\ref{eq:aggr} as \textit{neighbor aggregation}, and regard Equation~\ref{eq:update} as \textit{node update}.
The original GCN \cite{kipf2016semi} uses weighted summation for $\zeta^{(l)}$ and a single layer perceptron $\sigma\left(W^{(l)}z^{(l)}\right)$ for $\phi^{(l)}$ where $\sigma$ is a non-linear activation function.
Each layer in GCNs can be presented in a matrix form. For a given graph $\mathcal{G}=(\mathcal{V},\mathcal{E})$ with an adjacency matrix $A$, we define the propagation matrix $\widehat{A}$ as $\widehat{A}=\tilde{D}^{-1/2}\tilde{A}\tilde{D}^{-1/2}$, where $\tilde{A}=A+I$ and $\tilde{D}_{u,u}=\sum_v\tilde{A}_{u,v}$. We can write a GCN layer as:
\begin{align}
\left[H^{(l+1)}\right]^\top=\sigma\left(\widehat{A}\left[H^{(l)}\right]^\top\left[W^{(l)}\right]^\top\right) \label{eq:gcn}
\end{align}

\subsection{Partition Parallelism for GCN Training}
\label{sec:background_pp}
To improve the scalability of GCN training, many recent works follow the paradigm of partition parallelism~\cite{zhang2023survey}, which is depicted in 
 Figure~\ref{fig:gcn_comparison_b}. 
They either develop a scheduling algorithm towards balanced workload or optimized communication \cite{ma2019neugraph,zhu2019aligraph,jia2020improving,wang2021flexgraph,md2021distgnn,wang2022neutronstar,huang2024wisegraph} or adjust training algorithm to reduce or hide its communication overhead \cite{thorpe2021dorylus,wan2022pipegcn,ramezani2021learn,wan2022bns,peng2022sancus,chai2022distributed,bai2023staleness,wan2023adaptive,zhang2024sylvie}. 
However, as we will point out in Section~\ref{sec:mop_pp}, although partition-parallel training distributes the storage of \textit{giant feature tensors}, it suffers from scaled-out memory overhead and communication volume due to the duplicated remote neighbors (i.e., the red nodes in Figure~\ref{fig:gcn_comparison_b}). Furthermore, since a real-world graph is often highly irregular, finding a balanced-workload partition is NP-hard.

In parallel, CAGNET~\cite{tripathy2020reducing} and $P^3$~\cite{gandhi2021p3} explore the benefits of feature-level parallelism, but still lack either practicality or scalability. 
Specifically, CAGNET~\cite{tripathy2020reducing} splits node features along the feature dimension which are broadcasted to all devices during training, resulting in significantly redundant communication; $P^3$ \cite{gandhi2021p3} targets scalable distributed GCN training, but is still limited because it impractically assumes that the dimension of intermediate features is remarkably smaller than that of the input features~\cite{huang2024wisegraph}.

\subsection{Tensor Parallel Computing} 
Scalable training has been extensively studied for DNN models. For example, Horovod \cite{sergeev2018horovod}, PyTorchDDP \cite{li2020pytorch}, AutoDist \cite{zhang2020autosync}, BytePS \cite{jiang2020unified}, ZeRO \cite{rajbhandari2020zero}, and PyTorch-FSDP~\cite{zhao2023pytorch} leverage data parallelism for distributing independent input feature storage and the associated computation by duplicating the model parameters. In parallel, ColocRL \cite{mirhoseini2017device}, Mesh-Tensorflow \cite{shazeer2018mesh}, GPipe \cite{huang2019gpipe}, PipeDream \cite{harlap2018pipedream,narayanan2019pipedream}, Tofu \cite{wang2019supporting}, GSPMD \cite{xu2021gspmd}, TeraPipe \cite{li2021terapipe}, and GraphPipe~\cite{jeon2024graphpipe} develop model parallelism for storing model parameters distributedly, while depending on inter-model communication; FlexFlow \cite{lu2017flexflow}, Megatron-LM \cite{shoeybi2019megatron}, DeepSpeed~\cite{rasley2020deepspeed}, Alpa \cite{zheng2022alpa}, and Pathways \cite{barham2022pathways} combine both the above two parallelism to marry the best of both worlds.
However, although these systems have shown promising performance for scalable DNN training and even provided automated scheduling toolboxes, they are only applicable to dense tensor operations and thus do not work well for the scalable computing of GCNs (see Equation~\ref{eq:gcn}).

\begin{table}[t]
  \centering
  \caption{Summary of contribution in \MixGCN.}
  \label{tab:mixgcn_contribution}
  \begin{tabular}{l|c}
    \toprule
    \makecell[c]{\textbf{Challenge}} & \makecell[c]{\textbf{Innovation}} \\
    \midrule
    1. Giant Feature Tensors & Mixture of Parallelism (\S\ref{sec:mop}) \\
    \midrule
    \makecell[l]{2. Hybrid Sparse-Dense \\\hspace{7pt} Operations} & \makecell[c]{Mixture of Accelerators \\ (\S\ref{sec:moa})} \\
    \midrule
    \hspace{7pt} \makecell[l]{2.1 A Unique Sparse \\ \hspace{12pt} Operation: S-SpMM} & \makecell[c]{An Accelerator for \\ Operator Fusion (\S\ref{sec:moa_architecture})} \\
    \midrule
    \hspace{7pt} \makecell[l]{2.2 Unscalable Fine \\ \hspace{12pt} -grain Pipeline} & \makecell[c]{A Pipeline Scheduler with \\ Node Reordering (\S\ref{sec:moa_pipeline})} \\
    \bottomrule
  \end{tabular}
\end{table}

\subsection{GCN Accelerators}
\label{sec:background_acc}
To achieve aggressive efficiency improvement, dedicated accelerators for GCNs are highly desired. 
HyGCN \cite{yan2020hygcn}, GRIP \cite{kiningham2020grip}, G-CoS \cite{zhang2021g}, DyGNN \cite{chen2021dygnn}, GCoD \cite{you2021gcod}, OMEGA \cite{garg2021understanding}, Auten et al. \cite{auten2020hardware}, and Zhang et al. \cite{zhang2020hardware} develop heterogeneous accelerators to leverage the advantages of both dense and sparse accelerators. 
EnGN \cite{liang2020engn}, GCNAX \cite{li2021gcnax}, AWB-GCN \cite{geng2020awb}, and I-GCN \cite{geng2021gcn} follow a parallel direction by proposing reconfigurable architectures to optimize the computation of GCNs' hybrid operations.
Although these works have developed promising GCN inference accelerators, they cannot be directly utilized as our \MoA system's sparse accelerator due to their lack of support for both the output sparsities and scalable fine-grain pipeline (see Section~\ref{sec:moa} for details). In parallel,
GraphACT \cite{zeng2020graphact} and Rubik \cite{chen2021rubik} are the pioneering works for accelerating GCN training. 
Different from these training accelerators which mainly focus on architecture-level optimization, \MixGCN targets efficient and scalable GCN training by simultaneously integrating system- and architecture-level innovations.

\section{The Proposed Framework}
To address the two key bottlenecks in scalable GCN training -- \textit{giant feature tensors} and \textit{hybrid sparse-dense operations} -- we propose \MixGCN, which integrates \MoP and \MoA. During the implementation of \MoA, we identify that sparse operations in neighbor aggregation, coupled with an unscalable fine-grain pipeline, pose significant challenges to scalability. To overcome these challenges, we develop a dedicated accelerator that utilizes operator fusion and introduce a pipeline scheduler with node reordering to enhance efficiency. A summary of the innovations in \MixGCN is shown in Table 1.

\subsection{Mixture of Parallelism (MoP)}
\label{sec:mop}

\begin{figure*}
    \centering
    \includegraphics[width=1\linewidth]{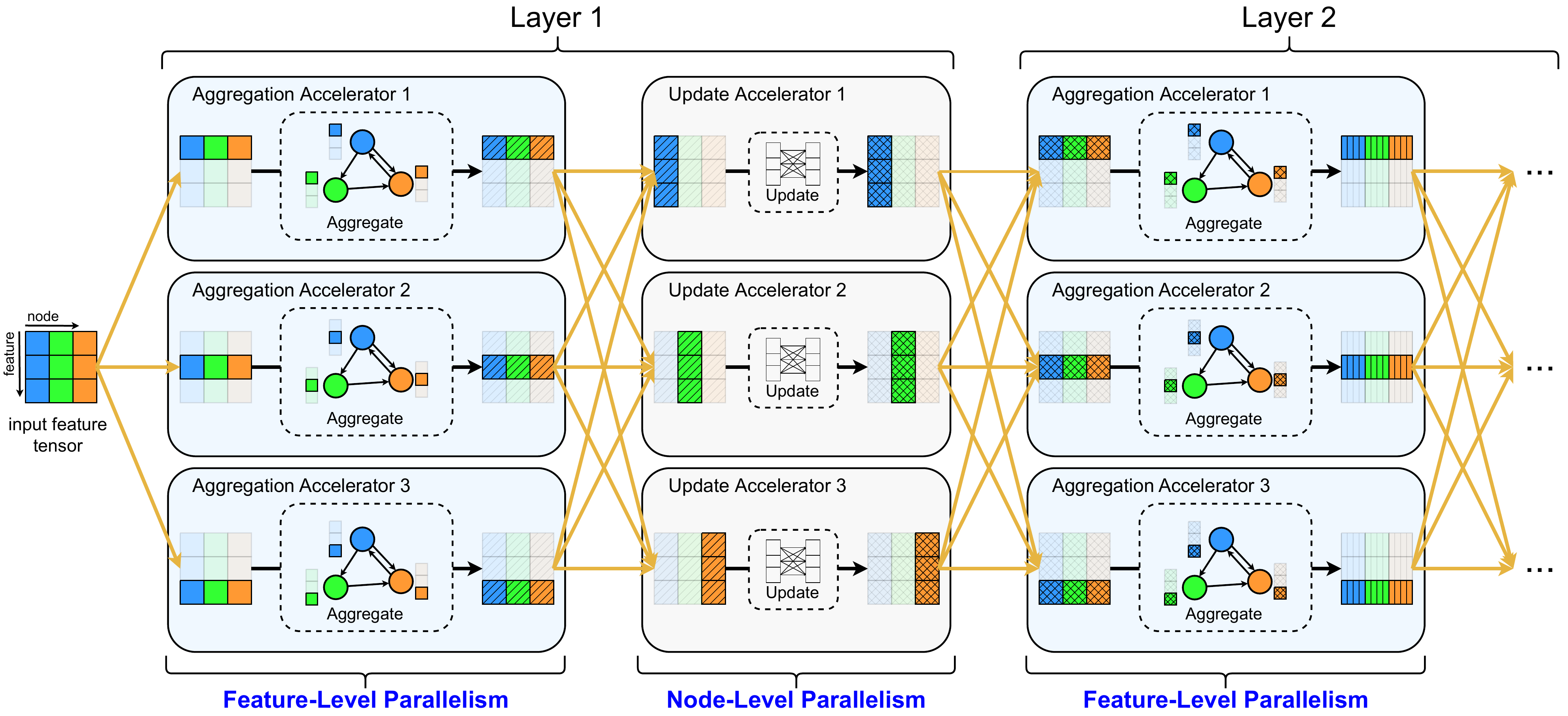}
    \caption{Illustrating the workflow of our proposed \textit{mixture of parallelism} (\textit{MoP}) where we adopt 3 pairs of aggregation and update accelerators for visual clarity.} 
    \label{fig:mixgcn_workflow}
\end{figure*}

{
\makeatletter
\newcommand{\removelatexerror}{\let\@latex@error\@gobble}
\makeatother
\begin{figure}[t]
\removelatexerror
\SetAlCapHSkip{0em}
\makeatletter
\patchcmd{\@algocf@start}
  {-1.5em}
  {0pt}
  {}{}
\makeatother

\begin{minipage}{0.478\textwidth}
\centering
\begin{algorithm}[H]
\SetAlgoLined
\SetInd{0.4em}{0.7em}
\Indmm\Indmm
\KwIn{adjacency matrix $\widehat{A}_i$, node feature $X_i$, label $Y_i$, initial model weight $W_0$}
\KwOut{trained model weight $W_T$ after $T$ iterations}
\Indpp\Indpp
$H^{(0)}_{in}\leftarrow X_i$\;
\For{$t\leftarrow0:T-1$}{
  \For{$l\leftarrow0:L-1$}{
  	Receiving remote features $H_{bd}^{(l)}$\label{line:comm}\;
  	$\left[H^{(l+1)}_{in}\right]^\top\leftarrow \sigma\left(\widehat{A}_i\left[H^{(l)}_{in}\,\middle\|\,H_{bd}^{(l)}\right]^\top\left[W_t^{(l)}\right]^\top\right)$\label{line:comp}\;
  }
  Estimate label $\widehat{Y}_i$ from $H^{(L)}_i$ and calculate $Loss(Y_i,\widehat{Y}_i)$\;
  Perform backward prop\label{line:bp1} and Update weight $W_{t+1}$\label{line:allreduce1}\;
}
\Return $W_T$
\caption{Partition parallelism for GCN training.}
\label{alg:pp}
\end{algorithm}
\end{minipage}
\vspace{-1em}
\end{figure}
}

\subsubsection{Partition Parallelism}
\label{sec:mop_pp}

An overview of partition parallelism is shown in Figure~\ref{fig:gcn_comparison_b}, and its detailed workflow is illustrated in Algorithm~\ref{alg:pp}. 
Specifically, each worker maintains a unique subgraph of the original graph by storing both the sub-adjacency matrix $\widehat{A}_i$ as defined in Equation~\ref{eq:gcn} and the input feature matrix $X_i$ of each node; For the computation of the $l$-th layer, each worker first collects dependent remote node features $H_{bd}^{(l)}$ from remote workers (line~\ref{line:comm}), and then calculates the features of inner nodes $H_{in}^{(l+1)}$ for the next layer (line~\ref{line:comp}). The backward propagation (line~\ref{line:bp1}) follows a similar paradigm by transferring the feature gradients of remote nodes. The model weights are synchronized via AllReduce (line~\ref{line:allreduce1}). 
Although many recent works on GCN training follow this paradigm as introduced in Section~\ref{sec:background_pp}, partition parallelism suffers from imbalanced computation according to the proposition below (see full proof in the supplementary material).

\begin{proposition}\label{prop:pp_balance}
Balancing the computation workload of GCN training with partition parallelism is NP-Hard.
\end{proposition}

In addition, \cite{wan2022bns} shows that the communication volume and feature memory requirement are linearly related to the total number of remote neighbors: 

\begin{proposition}\label{prop:pp_memory}
Communication volume and feature memory usage for partition parallelism are $\mathcal{O}(|\mathcal{R}|)$, where $\mathcal{R}$ is the set of remote neighbors.
\end{proposition}

{
\makeatletter
\newcommand{\removelatexerror}{\let\@latex@error\@gobble}
\makeatother
\begin{figure}[t]
\removelatexerror
\SetAlCapHSkip{0em}
\makeatletter
\patchcmd{\@algocf@start}
  {-1.5em}
  {0pt}
  {}{}
\makeatother

\begin{minipage}{0.478\textwidth}
\centering
\begin{algorithm}[H]
\SetAlgoLined
\SetInd{0.4em}{0.7em}
\Indmm\Indmm
\KwIn{adjacency matrix $\widehat{A}$, node feature $X$, label $Y$, initial model weight $W_0$}
\KwOut{trained model weight $W_T$ after $T$ iterations}
\Indpp\Indpp
$H^{(0)}\leftarrow X$\;
Set $d^{(l)}$ as the dimension of $l$-th layer\;
Set $k$ as the number of nodes\;
\For{$t\leftarrow0:T-1$}{
  \For{$l\leftarrow0:L-1$}{
    \ForAll(\Comment*[f]{aggregation in parallel}){worker$_\text{aggr}$ $i\in[m]$}{
        $p\leftarrow \left\lfloor\frac{d^{(l)}i}{m}\right\rfloor, q\leftarrow\left\lfloor\frac{d^{(l)}(i+1)}{m}\right\rfloor$\Comment*[r]{start/end index}
  	    $\left[Z^{(l)}_{p:q-1,:}\right]^\top\leftarrow\widehat{A}_i\left[H_{p:q-1,:}^{(l)}\right]^\top$\label{line:aggr}\;
  	}
    \ForAll(\Comment*[f]{update in parallel}){worker$_\text{upd}$ $i\in[m]$}{
        $p\leftarrow \left\lfloor\frac{ki}{m}\right\rfloor, q\leftarrow\left\lfloor\frac{k(i+1)}{m}\right\rfloor$\Comment*[r]{start/end index}
  	    $H^{(l+1)}_{:,p:q-1}\leftarrow\sigma\left(W_t^{(l)}Z_{:,p:q-1}^{(l)}\right)$\label{line:upd}\;
  	}
  }
  Estimate label $\widehat{Y}$ from $H^{(L)}$ and calculate $Loss(Y,\widehat{Y})$\label{line:loss}\;
  Perform backward prop\label{line:bp2} and Update weight $W_{t+1}$\;
}
\Return $W_T$
\caption{Mixture of parallelism for GCN training.}
\label{alg:mop}
\end{algorithm}
\end{minipage}
\vspace{-1em}
\end{figure}
}

\subsubsection{The Proposed Mixture of Parallelism (MoP)}
\label{sec:mop_mop}

The detailed workflow of \MoP is illustrated in Figure~\ref{fig:mixgcn_workflow} and described in Algorithm~\ref{alg:mop}. For a given input feature matrix $H^{\left(0\right)}$, \MoP first splits $H^{\left(0\right)}$ along its feature dimension, and distributes each split to different accelerators for \textit{neighbor aggregation} (see the left part of Figure~\ref{fig:mixgcn_workflow}), each of which computes the corresponding features (rows) of $A\left[H^{\left(0\right)}\right]^\top$ by multiplying $A$ and the assigned inputs (line~\ref{line:aggr} of Algorithm~\ref{alg:mop}). Next, \MoP performs all-to-all communication so that each accelerator for \textit{node update} gets the access to the entire features of assigned nodes, and updates them by the stored model weights (line~\ref{line:upd}). Finally, all-to-all communication is performed again, enabling all the aggregation accelerators to perform computation for the next layer. \MoP repeats this process until the last layer. The backward pass of \MoP follows a similar workflow (line~\ref{line:bp2}). 
It is worth noting that in Algorithm~\ref{alg:mop}, line~\ref{line:loss} is also completed in a distributed manner but not shown.

Contrary to partiton parallelism, balancing the workload of the aggregation accelerators or update accelerators is trivial with \MoP, because uniformly splitting the feature tensor naturally guarantees a fully balanced workload across both the aggregation or update accelerators. Therefore, we have the following statements.

\begin{proposition}\label{prop:mop_balance}
Balancing the computation of GCN training with \MoP can be solved in $\mathcal{O}(1)$ time.
\end{proposition}

Furthermore, the node features are never replicated during the process of data transfer in \MoP, which ensures constant communication volume and feature memory consumption.

\begin{proposition}\label{prop:mop_memory}
Both the communication volume and feature memory requirement are $\mathcal{O}(\mathcal{N})$ for GCN training with \MoP, where $\mathcal{N}$ is the size of node set.
\end{proposition}

\niparagraph{Remark} Based on the above discussion of partition parallelism and \MoP, we summarize their differences below:

\begin{itemize}
\item \textbf{Computation Workload.} According to Proposition \ref{prop:pp_balance} and Proposition~\ref{prop:mop_balance}, since balancing computation workload for partition parallelism is impossible, \MoP enjoys better scalability for its strictly balanced workload.
\item \textbf{Communication Volume and Feature Memory Requirement.} With the increasing number of accelerators, the remote neighbor set is growing. As a result, partition parallelism suffers from scaled-out communication volume and feature memory requirement according to Proposition~\ref{prop:pp_memory}. On the other hand, Proposition~\ref{prop:mop_memory} ensures constant communication volume and feature memory usage, regardless of the number of accelerators. 
Therefore, \MoP offers better scalability.
\item \textbf{All-to-All Communication.} Both partition parallelism and \MoP employ all-to-all communication. Nevertheless, in addition to the required constant communication volume as mentioned above, \MoP enjoys a more regular communication pattern, securing balanced communication workload and better scalability.
\end{itemize}

\begin{figure}[t]
    \centering
    \includegraphics[width=1\linewidth]{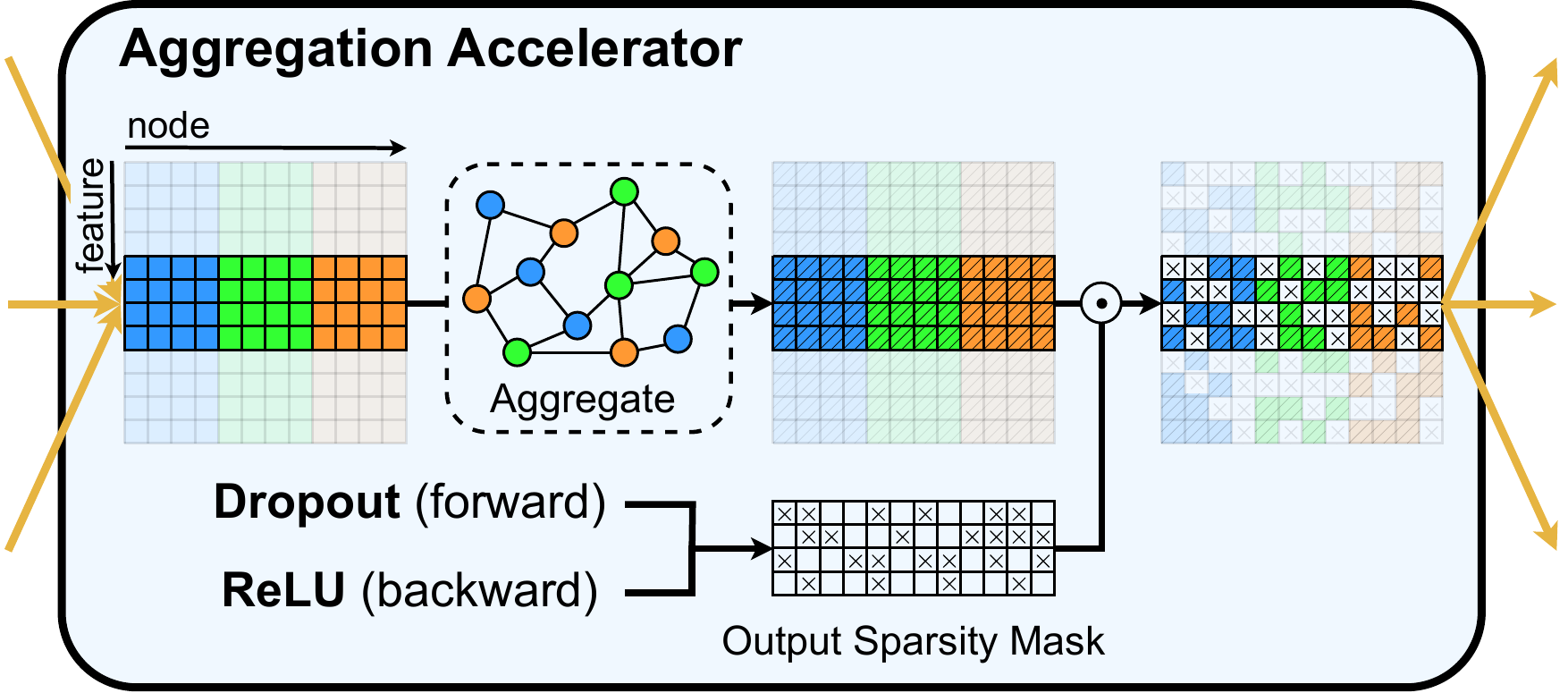}
    \caption{An illustration of S-SpMM in the accelerator for \textit{neighbor aggregation}.}
    \label{fig:output_sparsity}
\end{figure}

\subsubsection{Scalability of All-to-All Communication}
\label{sec:mop_alltoall}
One potential concern for adopting \MoP is that it relies on all-to-all communication which may restrict its scalability. Despite that all-to-all communication also exists in partition parallelism and requires a more irregular communication pattern than \MoP, we justify that under a proper design, all-to-all communication is not costly. For example, \cite{yang2000optimal,massini2003all} leverage a butterfly topology to implement all-to-all communication by assuming that arbitrary lengths of wires are acceptable, which only requires $\mathcal{O}(n\log n)$ wires and $\mathcal{O}(\log n)$ stages for non-blocking data transfer where $n$ is the number of accelerators. In practice, all-to-all communication has been widely adopted in large-scale Transformer training and inference \cite{lepikhin2020gshard,fedus2021switch,rajbhandari2022deepspeed} for connecting 
up to 2048 devices. Our experiments also verify its scalability (see Figure~\ref{fig:breakdown}).

\begin{figure*}[ht]
    \centering
    \includegraphics[width=1\linewidth]{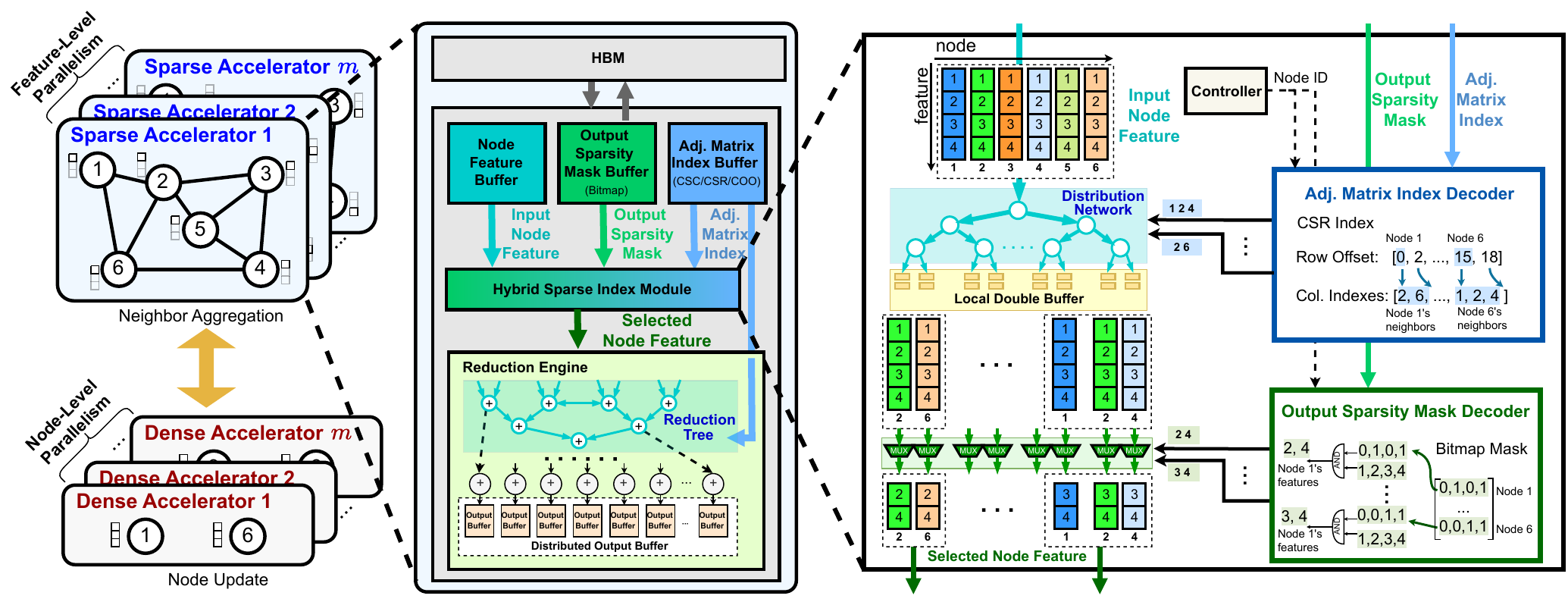}
    \captionsetup[subfloat]{captionskip=-1em}
    \subfloat[\label{fig:MoA_arch-a} The framework of \MoA.]{\hspace{.25\linewidth}}
    \subfloat[\label{fig:MoA_arch-b} The architecture of our accelerator.]{\hspace{.26\linewidth}}
    \subfloat[\label{fig:MoA_arch-c} The hybrid sparse index module.]{\hspace{.49\linewidth}}
    \caption{An illustration of the proposed \emph{mixture of accelerators} (\MoA), which integrates a dedicated accelerator for computing S-SpMM (Sampled Sparse Matrix-Matrix Multiplication).}
    \label{fig:MoA_arch}
\end{figure*}

\subsection{Mixture of Accelerators (MoA)}
\label{sec:moa}

\label{sec:moa_intro}
Benefiting from the proposed \MoP, the second challenge, \textit{hybrid sparse-dense operations}, 
can be naturally resolved by assigning the sparse and dense operations to two different groups of accelerators (outlined in Figure~\ref{fig:MoA_arch-a}). 
We propose \fullMoA (\MoA) on top of our \MoP to leverage (a group of identical) sparse accelerators to accelerate the sparse matrix operations (i.e., \textit{neighbor aggregation}) and (another group of identical) dense accelerators to accelerate the dense matrix operations (i.e., \textit{node update}).

\subsubsection{An Accelerator for Operator Fusion}
\label{sec:moa_architecture}

\niparagraph{Motivation} As discussed in Section~\ref{sec:background_acc}, existing GCN inference accelerators are not optimal for accelerating GCN training. 
This stems from the unique sparse operation -- S-SpMM -- an operation that has yet to be thoroughly studied. Specifically, as shown in Figure \ref{fig:output_sparsity}, the forward and backward passes of the aggregation computation in GCN training can be formulated as follows: 
\begin{align} 
Z=AH\odot M \label{eq:fusion}
\end{align}
where $H$ is the feature matrix, $\odot$ denotes element-wise product, $A$ represents the propagation matrix, and $M$ is a sparse mask matrix.
In most GCN training tasks, the mask matrix $M$ drops about 50\% of the outputs after \emph{neighbor aggregation} due to either dropout (forward) or ReLU (backward). This mask presents an opportunity to reduce both computation and data movement during training, motivating us to design an accelerator by fusing these two sparse operations.

We define the fused operation in Equation~\ref{eq:fusion} as S-SpMM (Sampled Sparse Matrix-Matrix Multiplication), which extends two traditional sparse operations: SpMM and SDDMM.

\begin{figure*}[t]
\centering
    \includegraphics[width=1\linewidth]{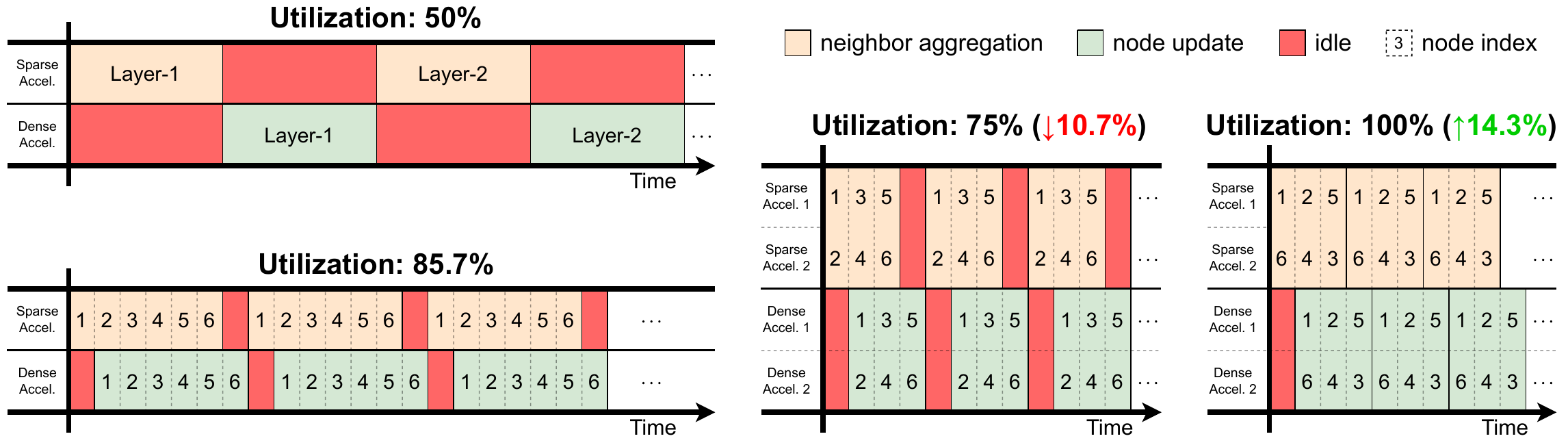}

    \captionsetup[subfloat]{captionskip=-1em,margin={0cm,9.2cm}}
    \vspace{-8.0em}
    \subfloat[Naive workflow.]{
        \hspace{0.95\linewidth}
        \label{fig:pipeline_a}
    }
    
    \vspace{6.7em}

    \clearcaptionsetup[margin]{subfloat}

    \subfloat[Pipeline in a single pair of accelerators.]{
        \hspace{.47\linewidth}
        \label{fig:pipeline_b}
    }\hfill
    \subfloat[Pipeline in a scalable system.]{
        \hspace{.25\linewidth}
        \label{fig:pipeline_c}
    }\hfill
    \subfloat[Pipeline with node reordering.]{
        \hspace{.25\linewidth}
        \label{fig:pipeline_d}
    }

\caption{An example that illustrates the comparison of the temporal execution flow among different pipeline designs between the sparse and dense accelerators. We assume that the training graph is identical to the graph in Figure~\ref{fig:gcn_comparison_a}.}
\label{fig:pipeline_comparison}
\end{figure*}

\niparagraph{Architecture Overview}
Figure \ref{fig:MoA_arch-b} illustrates the overall architecture to compute S-SpMM, consisting of buffers (top), a hybrid sparse index module with a distribution network (middle), and a reduction engine with a reduction network (bottom). 
To coordinate the graph sparsity and the unique output sparsities, we introduce (1) an adjacency matrix index buffer and an output mask buffer (Figure \ref{fig:MoA_arch-b} (top))  to store the corresponding sparsity indexes using proper index formats, e.g., CSR (compressed sparse row) format for representing the adjacency matrix of the graph and bitmap format for the output sparsities \cite{dave2021hardware} besides the node buffer to store the input nodes from the corresponding dense accelerators; (2) the hybrid sparse index module (Figure \ref{fig:MoA_arch-b} (middle)) which takes the adjacency matrix index and the node feature masks from the corresponding two buffers and then use them to select the required neighbors' features from the node buffer to avoid the unnecessary computation of the reduction engine; and (3) the reduction engine (Figure \ref{fig:MoA_arch-b} (bottom)) for sparse \textit{neighbor aggregation} operations. 
We provide more details about the key module, i.e., the  hybrid sparse index module, in the following. 

\niparagraph{The Hybrid Index Module}
Figure~\ref{fig:MoA_arch-c} shows the proposed hybrid index module, 
where the two steps of selections and decoders are involved to enable a two-step data selection: a node-wise selection based on the graph sparsity by using the distribution network and a feature-wise selection based on the output sparsities by using the multiplexers (MUXs). 
In particular, at the first step, the adjacency matrix index decoder takes the index from both the adjacency matrix index buffer and the node IDs from the controller to identify each node's neighbor nodes and then uses the fat tree-based distribution network~\cite{kwon2018maeri} to feed the neighbor nodes to the local double buffers, e.g., selecting both node 2 and node 6 for aggregating the neighbors of node 1 in Figure~\ref{fig:MoA_arch-c} and feeding them to the local double buffers.

In the second step, the output mask decoder selects the required neighbors' features from the local double buffers to save the data movements to the reduction engine and the computations. The required neighbors' feature selection is achieved by simply performing ``AND'' between the features and the corresponding feature masks in the MUXs.
Thanks to this proposed hybrid sparse index module, only the required neighbor node features need to be selected from the node buffer to be executed in the reduction engine for \textit{neighbor aggregation} operations. 
In addition, the two steps of selection and the following reduction can be pipelined by adopting the local double buffers and properly designing the sparsities and the distribution network bandwidth.

\subsubsection{A Pipeline Scheduler with Node Reordering}
\label{sec:moa_pipeline}

\niparagraph{Motivation}
As shown in Figure~\ref{fig:pipeline_a}, the naive workflow of \MoA suffers from low hardware utilization. A common approach to alleviate this is adopting a fine-grain pipeline \cite{geng2020awb}, as depicted in Figure~\ref{fig:pipeline_b}. However, this can still result in frequent idle periods, because the sparse accelerator cannot start processing the next operations until all dependent dense operations are complete. 
For example, in the graph shown in Figure~\ref{fig:gcn_comparison_a}, node 1 depends on node 6, leading to idle time before node 1 can be processed. This idleness is further exacerbated in scalable training scenarios, where the workload per accelerator decreases, but the minimum granularity required for full hardware utilization remains unchanged (see Figure~\ref{fig:pipeline_c}).

\begin{figure}[t]
  \centering
  \subfloat[The workflow of a naive pipeline.]{
        \includegraphics[width=0.99\linewidth]{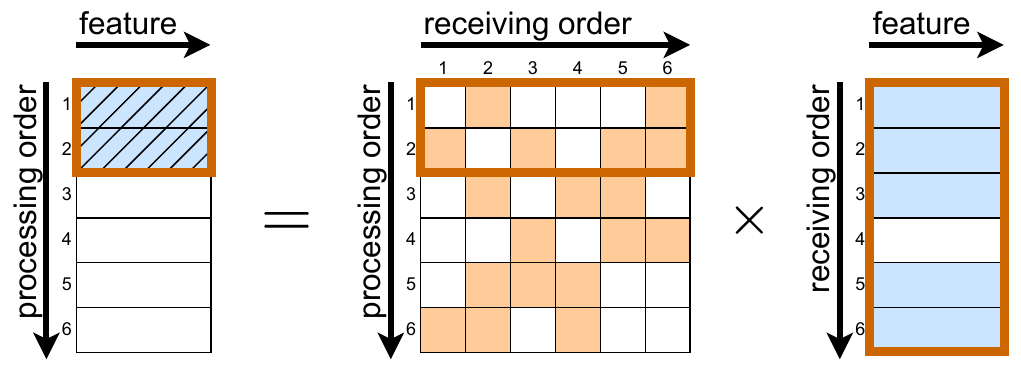}
        \label{fig:moa_node_order_a}
    }
    
    \subfloat[The workflow with node reordering.]{
        \includegraphics[width=0.99\linewidth]{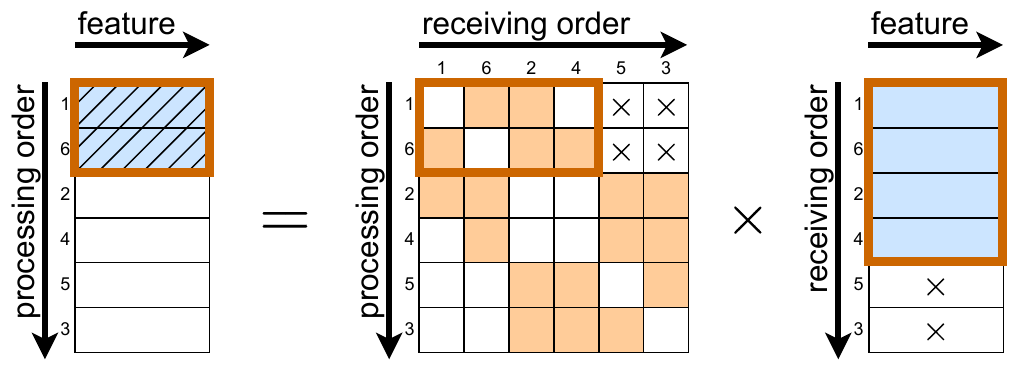}
        \label{fig:moa_node_order_b}
    }
\caption{An illustrative comparison between the existing pipeline and our optimized approach, using the example of processing/receiving two nodes per pipeline step. We assume the training graph is the same as the one shown in Figure~\ref{fig:gcn_comparison_a}.}
\label{fig:moa_node_order}
\end{figure}

This idle time can be reduced by leveraging node reordering. As illustrated in Figure~\ref{fig:moa_node_order_a}, the first processing batch in the original schedule depends on 5 out of 6 nodes. 
In contrast, with node reordering in the optimized schedule (see Figure~\ref{fig:moa_node_order_b}), the first processing batch only require the first 4 nodes (i.e., the first two processing batches), enabling the pipeline to eliminate idle periods, as shown in Figure~\ref{fig:pipeline_d}.

\begin{figure*}[t]
    \centering
    \includegraphics[width=1\linewidth]{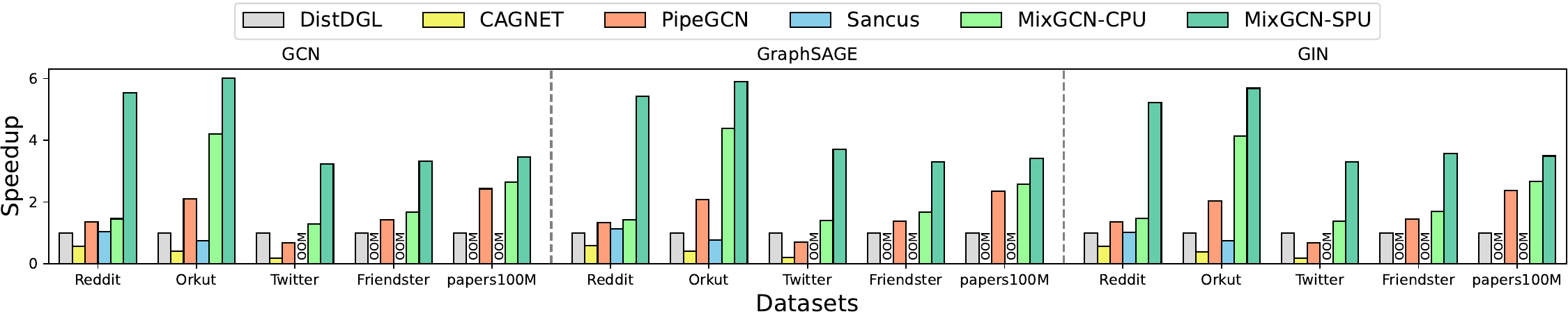}
    \caption{Speedup comparison between \MixGCN and baselines against DistDGL on 4 nodes.}
    \label{fig:throughput}
\end{figure*}

\niparagraph{Method} 
Building upon this motivation, the core strategy to reduce idleness in \MoA is to eliminate dependencies between the first and last processing batches. Our goal is to determine an optimized node ordering that accomplishes this.

Formally, let the position of node $v$ in the processing sequence be denoted as $p_v$. We define $b \triangleq \max\limits_{(u,v)\in\mathcal{E}}\{p_v - p_u\}$, where $\mathcal{E}$ represents the edge set. This ensures that for any node $u$, each of its dependent neighbors $v$ satisfies $p_v - p_u \leq b$. By optimizing the value of $b$, we minimize the maximum latency for gathering all neighbor features.

In graph theory, the value $b$, as defined above, corresponds to the \textit{graph bandwidth} \cite{chinn1982bandwidth}, which can be optimized using the reverse Cuthill–McKee algorithm in $\mathcal{O(|E|)}$ time \cite{cuthill1969reducing}. This complexity is identical to that of METIS \cite{karypis1998fast}, a widely-used algorithm for graph partitioning in partition parallelism. Since the input graph of a GCN is always symmetric~\cite{kipf2016semi}, we emphasize that the optimal node ordering can also be reused for backward propagation, thereby amortizing preprocessing overhead.

\niparagraph{Remark} We offer a method to quantify the scalability of a pipeline schedule. For a graph with $n$ nodes and bandwidth $b$, when the processing nodes are divided into $s$ batches, idleness can be eliminated if all nodes in the first batch depend only on nodes within the first $s-1$ batches. Mathematically, this condition can be expressed as $\frac{(s-1)n}{s} - \frac{n}{s} \leq b$. Based on this, we present the following proposition: 
\begin{proposition}
\label{prop:utilize}
The idleness in a fine-grain pipeline can be eliminated when $\frac{n-b}{2n} \geq \frac{1}{s}$. 
\end{proposition}

\begin{table}[t]

\setlength{\tabcolsep}{0.39em}
\centering

\caption{Details of the five large-scale graph datasets.}
\label{tab:dataset}

\begin{tabular}{c|ccccc}
\toprule
Dataset & \# Nodes & \# Edges & \# Feats & \# Classes \\
\midrule
Reddit & 233K & 115M & 602 & 41  \\
Orkut & 3.07M & 117M & 320 & 20 \\
Twitter & 41.7M & 1.47B & 52 & 16 \\
Friendster & 65.6M & 1.81B & 128 & 64  \\
ogbn-papers100M & 111M & 1.62B & 128 & 172  \\
\bottomrule
\end{tabular}
\end{table}

As such, we define the \emph{minimum granularity} of a pipeline as $\frac{n-b}{2n}$, which measures the scalability of a pipeline schedule. According to this claim, a moderately large $b$ (e.g., $0.8n$) is sufficient for scalable GCN training, and a smaller $b$ has better scalability by allowing more accelerators with a smaller $s$.

\section{Experiments}
\label{sec:experiment}

\subsection{Experimental Setting}
\niparagraph{The Model and Datasets} We evaluate the performance of \MixGCN on three popular architectures: GCN~\cite{kipf2016semi}, GraphSAGE~\cite{hamilton2017inductive}, and GIN~\cite{xu2018powerful}. Each model consists of 3 layers with 128 hidden units and is trained on five large-scale datasets: Reddit~\cite{hamilton2017inductive}, Orkut~\cite{yang2012defining}, Twitter~\cite{kwak2010twitter}, Friendster~\cite{yang2012defining}, and ogbn-papers100M~\cite{hu2020open}. The details of these datasets are provided in Table~\ref{tab:dataset}.

\niparagraph{Implementation} \MixGCN is implemented in DGL \cite{wang2019dgl} and PyTorch \cite{li2020pytorch}. We set the default communication backend as NCCL and conduct the experiments on a 4-node cluster. Each computation node is equiped with 8 H100 GPUs and a 64-core Intel Xeon Platinum 8462Y+ CPU. The nodes are connected with InfiniBand.

\niparagraph{The Underlying Sparse Accelerators} Because concurrent distributed systems do not support our proposed \MoA, we consider two variants of \MixGCN with different underlying sparse accelerators. (1) To implement \MixGCN in a real system, we run \emph{node update} in GPUs while computing \emph{neighbor aggregation} in CPUs. Each worker under this setting is assigned 1 GPU and 8 cores of CPU for performing \textit{node update} and \textit{neighbor aggregation}, respectively. The system under this setting is dubbed as \MixGCN-CPU.
(2) We also evaluate the performance of \MoA with the proposed sparse accelerator through simulation. The dedicated accelerator is implemented with a commercial 28nm CMOS technology using Synopsys's Design Compiler for gate-level netlist \cite{DC} and the Memory Compilers from the foundry. The accelerator has 16,384 floating-point 32-bit adders and 100MB on-chip SRAM, including a 32MB node feature buffer, a 32MB distributed output buffer, a 32MB adjacency matrix index buffer, and a 4MB output sparsity mask buffer, resulting in an area of 283.35$mm^2$ and a power of 38.96W at 500MHz clock frequency. Since the adjacency matrix is binary, no multiplier is needed in the proposed sparse accelerator. The accelerator area is constrained by on-chip SRAM (i.e., 77.6\% of the overall area) which can be reduced by using advanced memories like e-DRAM in HyGCN~\cite{yan2020hygcn}.
For the HBM settings, we choose 1024 GB/s following the existing dense accelerator design (e.g., TPU \cite{jouppi2017datacenter} and GPU \cite{lindholm2008nvidia}). 
We name \MixGCN with the proposed sparse processing units as \MixGCN-SPU.

\begin{figure}[t]
    \centering
    \includegraphics[width=1.0\linewidth]{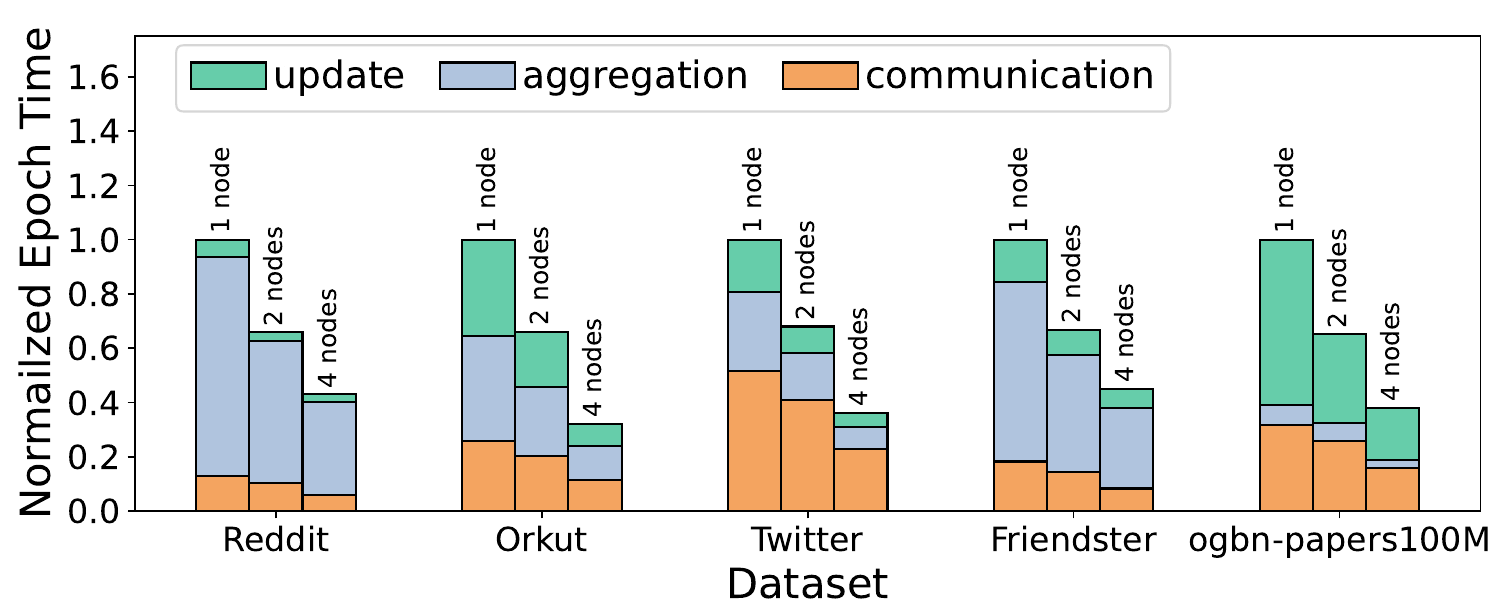}
    \caption{The time breakdown of \MixGCN-CPU without fine-grain pipeline.}
    \label{fig:breakdown}
\end{figure}

\begin{figure*}[t]
    \centering
    \includegraphics[width=1\linewidth]{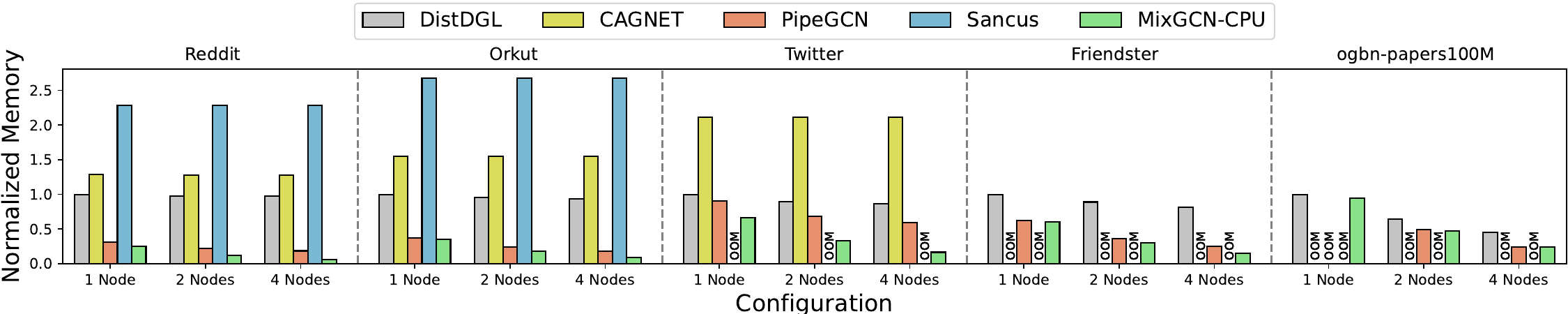}
    \caption{Normalized memory for GCN training between \MixGCN and baseline methods against DistDGL on a single node.}
    \label{fig:memory}
\end{figure*}

\begin{figure*}[t]
    \centering
    \includegraphics[width=1\linewidth]{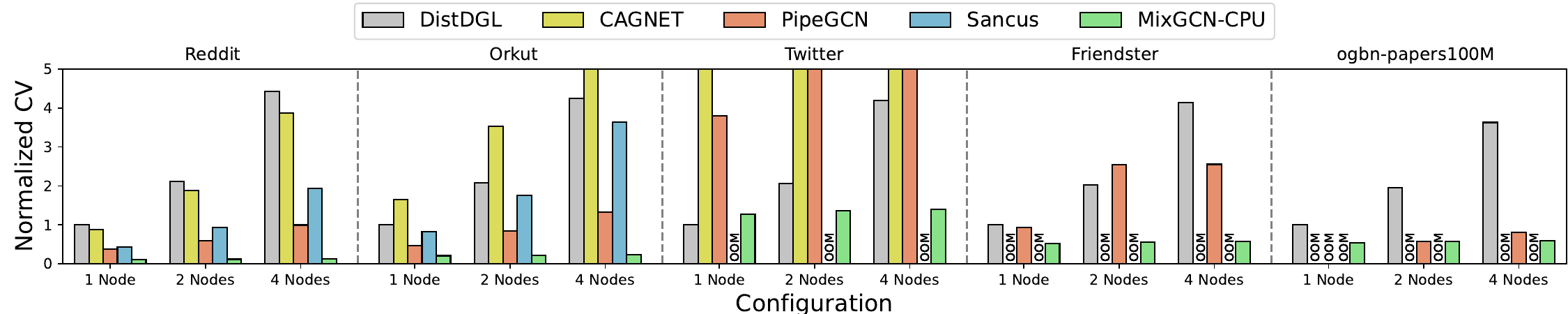}
    \caption{Normalized communication volume for GCN between \MixGCN and baselines against DistDGL on a single node.}
    \label{fig:cv}
\end{figure*}

\begin{figure*}[t]
    \centering
    \includegraphics[width=1\linewidth]{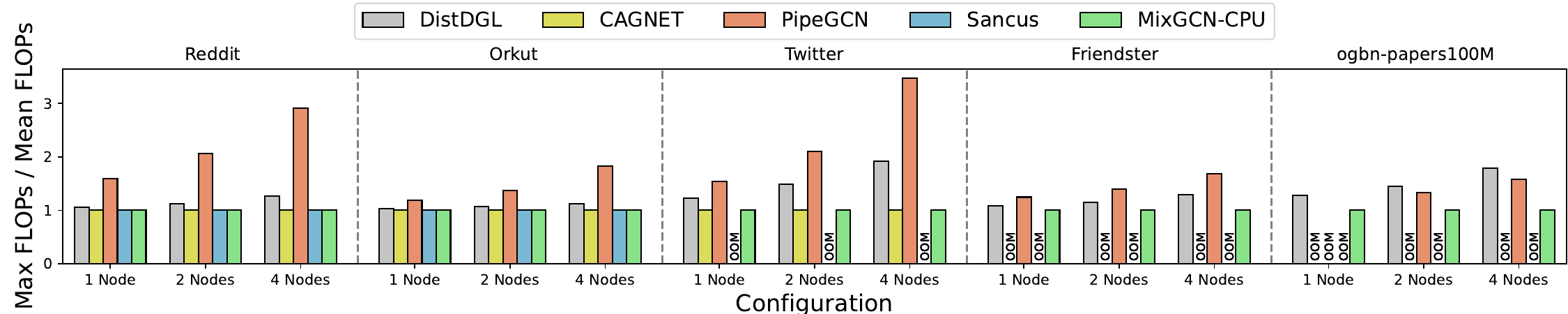}
    \caption{Workload balance comparison between \MixGCN and baselines.}
    \label{fig:workload}
\end{figure*}

\niparagraph{The Baseline Methods} To understand the end-to-end advantages of \MixGCN, we compare \MixGCN with DistDGL \cite{zheng2020distdgl}, 
CAGNET~\cite{tripathy2020reducing}, PipeGCN~\cite{wan2022pipegcn}, and Sancus~\cite{peng2022sancus}, which represent different variants of partition parallelism. We discuss their design differences in Section~\ref{sec:discussion-pp}.

\subsection{Comprehensive Performance}
\label{sec:exp_e2e}

\niparagraph{Overall Throughput Comparison} Figure~\ref{fig:throughput} presents a comparison of the throughput speedup between \MixGCN and various baseline systems, all measured against DistDGL in a 4-node setup. In particular, \MixGCN-CPU demonstrates significant performance gains, achieving up to 4.2$\times$, 10.4$\times$, 1.9$\times$, and 5.5$\times$ higher throughput compared to DistDGL, CAGNET, PipeGCN, and Sancus, respectively. 
Additionally, \MixGCN-SPU further enhances the efficiency of \MixGCN-CPU, delivering an additional speedup of up to 3.8$\times$. These results highlight the substantial improvements in both versions of \MixGCN over existing GCN systems.

\niparagraph{Time Breakdown} 
Figure~\ref{fig:breakdown} shows the time breakdown of \MixGCN-CPU. We disable the pipeline introduced in Section~\ref{sec:moa_pipeline} to analyze the exact time cost for each operation. 
We observe that \emph{neighbor aggregation} is the main bottleneck in Reddit, which is the cause of substantial improvement of \MixGCN-SPU over \MixGCN-CPU. Both communication and computation incur less overhead with increased workers because they are distributed across workers. 

\subsection{The Performance of \MoP}
\label{sec:exp_mop}

\niparagraph{Memory Usage} 
Figure~\ref{fig:memory} presents a normalized memory usage comparison for GCN training between \MixGCN and baseline methods against DistDGL on a single node. The results indicate that \MixGCN consistently utilizes less memory than baseline methods during GCN training. Notably, \MixGCN exhibits linear memory scaling with an increasing number of computation nodes, underscoring its strong scalability. In contrast, baseline methods display poor scalability, with only minimal memory savings as the number of workers increases. This experiment provides empirical evidence supporting Proposition~\ref{prop:pp_memory} and Proposition~\ref{prop:mop_memory} by highlighting the superior scalability of \MixGCN.

\niparagraph{Constant Communication Volume} Figure~\ref{fig:cv} presents a comparison of the normalized communication volume for GCN training between \MixGCN and baseline methods against DistDGL on a single node. The results demonstrate that \MixGCN maintains a constant communication volume, unaffected by the increase in computation nodes. In contrast, baseline models incur significantly heavier communication overhead as the number of computation nodes grows, leading to poor scalability. This stark difference in communication efficiency further validates Proposition~\ref{prop:pp_memory} and Proposition~\ref{prop:mop_memory}, underscoring the scalability advantages of \MixGCN.

\niparagraph{Balanced Workload} To verify the benefits of the balanced workload of \MixGCN, we measure the ratio between the maximum FLOPs and average FLOPs of all workers, and report the results in Figure~\ref{fig:workload}. DistDGL and PipeGCN suffer from imbalanced workloads when the computational nodes are increased because more partitions create more diverse subgraphs. CAGNET and Sancus have balanced computation because it adopts random partition, but incurs overwhelming communication overhead as shown in Figure~\ref{fig:cv}. \MixGCN creates a fully balanced workload, enabling scaling-up benefits. 

\begin{figure}[t]
    \centering
    \includegraphics[width=1\linewidth]{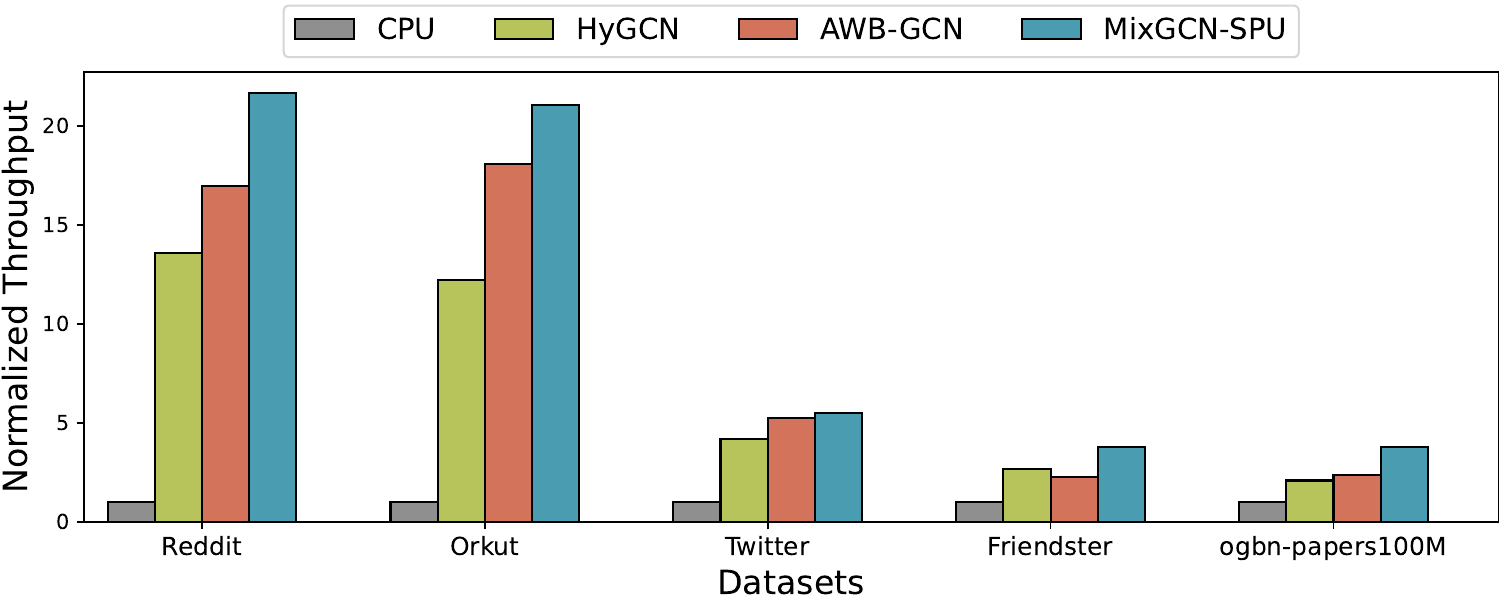}
    \caption{Speedup over CPU of the proposed sparse accelerator and the baseline accelerators: HyGCN and AWB-GCN.}
    \label{fig:speedup_sparse}
\end{figure}

\subsection{The Performance of \MoA}
\label{sec:exp_moa}


\begin{table}[t]
    \centering
    \caption{Energy consumption comparison of our proposed accelerator and baselines (units: J)}
    \label{tab:exp_energy}
    \setlength{\tabcolsep}{0.22em}
    \begin{tabular}{c|ccccc}
        \toprule
        Dataset & Reddit & Orkut & Twitter & Friendster & papers100M \\
        \midrule
        CPU     & 146.9k  & 164.1k & 3291k  & 3843k & 4322k \\
        HyGCN & 19.3 & 43.9 & 533.9 & 1031 & 1547\\
        AWB-GCN   & 6.9 & 33.6 & 2326 & 6250 & 6810 \\
        Ours  & 7.6 & 9.5 & 605.5 & 1135 & 1269 \\
        \bottomrule
    \end{tabular}
\end{table}


\subsubsection{The Dedicated Accelerator}

\niparagraph{Speedup of \MixGCN-SPU} Figure~\ref{fig:speedup_sparse} illustrates the performance gains of our proposed sparse accelerator compared to CPU, HyGCN~\cite{yan2020hygcn}, and AWB-GCN~\cite{geng2020awb} when executing \textit{neighbor aggregations} on the evaluated datasets. To ensure a fair comparison, we configured HyGCN and AWB-GCN with identical computational resources (16,384 units) and on-chip SRAM size (100MB), while setting a uniform HBM bandwidth of 1024GB/s for all three sparse accelerators. We maintained the original clock frequencies of the baseline accelerators: 1GHz for HyGCN and 330MHz for AWB-GCN. Our proposed sparse accelerator demonstrates significant performance improvements, achieving speedups of 3.77$\times$ to 21.6$\times$ over CPU, 1.32$\times$ to 1.80$\times$ over HyGCN, and 1.05$\times$ to 1.66$\times$ over AWB-GCN across the evaluated datasets.

\niparagraph{Power Consumption} We compared the energy consumption of our proposed accelerator against HyGCN and AWB-GCN, as well as CPU. The results are summarized in Table~\ref{tab:exp_energy}. In our experiments, the accelerator demonstrate up to 3.5$\times$ and 5.5$\times$ improvement over HyGCN and AWB-GCN, respectively. These results highlight the effectiveness of our accelerator in reducing energy consumption while maintaining performance across various graph datasets.

\subsubsection{The Fine-grain Pipeline}

\niparagraph{The Impact of Node Reordering} We investigate the performance benefits of node reordering in fine-grain pipelines for GCN training, as presented in Table~\ref{tab:exp_reorder}. According to Proposition~\ref{prop:utilize}, the number of pipeline stages is determined by $\lceil\frac{2n}{n-b}\rceil$. Our evaluation on a four-node setup reveals that node reordering yields a throughput improvement of up to 1.18$\times$ compared to pipelines without node reordering, confirming its effectiveness. However, for datasets like Reddit, where GCN training is hindered by significant latency imbalances between neighbor aggregation and node update, the benefits of node reordering are minimal, as evidenced by a modest 1.01$\times$ speedup.

\niparagraph{Preprocessing Overhead} We compare the preprocessing overhead of RCM adopted in \MixGCN for node reordering with METIS, a widely used algorithm in partition parallelism as adopted in DistDGL and PipeGCN. Table~\ref{tab:exp_preprocessing} presents the results. Although both METIS and RCM have the same time complexity $\mathcal{O}(\mathcal{E})$ and are implemented in C++, METIS incurs significant I/O overhead due to its hierarchical partitioning approach, which requires storing intermediate results to disk. In contrast, RCM is a straightforward variant of Breadth-First Search (BFS), resulting in substantially lower preprocessing overhead. As shown in Table~\ref{tab:exp_preprocessing}, RCM outperforms METIS by a considerable margin, with preprocessing times reduced by up to two orders of magnitude.


\begin{table}[t]
    \centering
    \caption{Speedup of node reordering in fine-grain pipelines for GCN training across four nodes.}
    \label{tab:exp_reorder}
    \setlength{\tabcolsep}{0.29em}
    \begin{tabular}{c|ccccc}
        \toprule
        Dataset & Reddit & Orkut & Twitter & Friendster & papers100M \\
        \midrule
        Stages & 11 & 11 & 19 & 6 & 4\\
        Speedup & 1.01$\times$ & 1.07$\times$ & 1.02$\times$ & 1.13$\times$ & 1.18$\times$ \\
        \bottomrule
    \end{tabular}
\end{table}



\begin{table}[t]
    \centering
    \caption{Comparison of preprocessing overhead for METIS and RCM (Units: second).}
    \label{tab:exp_preprocessing}
    \setlength{\tabcolsep}{0.32em}
    \begin{tabular}{c|ccccc}
        \toprule
        Dataset & Reddit & Orkut & Twitter & Friendster & papers100M \\
        \midrule
        METIS & 342 & 1009 & 23473 & 32490 & 13660 \\
        RCM & 4.25 & 9.39 & 175 & 209 & 201 \\
        \bottomrule
    \end{tabular}
\end{table}


\section{Discussion}

\subsection{Comparison with Related Works}\label{sec:discussion-pp}

\niparagraph{Systems with Partition Parallelism} As introduced in Section \ref{sec:background_pp}, partition parallelism is the predominant strategy employed in distributed GCN training. Beyond the standard algorithm outlined in Algorithm~\ref{alg:pp}, several variants exist. For instance, DistDGL~\cite{zheng2020distdgl}, in full-graph training, collects all $L$-hop neighbors (where $L$ is the layer number) before each training iteration, thereby avoiding inter-layer communication. PipeGCN~\cite{wan2022pipegcn} is a representative approach that leverages historical node features to facilitate asynchronous communication. Sancus~\cite{peng2022sancus} further optimizes communication by adopting intermittent data transfer and replacing peer-to-peer communication with broadcast, resulting in a more regular communication pattern.

However, as discussed in Section~\ref{sec:mop} and evaluated in Section \ref{sec:experiment}, systems based on partition parallelism face challenges such as scaling out total communication volume and feature memory, as well as suffering from imbalanced workloads. 
This work identifies \MoP as a promising approach due to its ability to maintain constant communication volume, scalable memory usage, and balanced workload.

\niparagraph{Systems with Feature-level Parallelism} Building upon partition parallelism, CAGNET~\cite{tripathy2020reducing} and $P^3$~\cite{gandhi2021p3} incorporate feature-level parallelism to reduce memory usage. CAGNET explores hybrid partitioning for the input feature tensor and adjacency matrix of a GCN layer. However, its communication strategy relies on broadcast communication, which incurs redundant communication due to not considering the sparse pattern of the adjacency matrix. In contrast, $P^3$ applies feature-level parallelism to the \textit{node update} process of the first layer, assuming the hidden dimension is significantly smaller than the input dimension. This approach distributes the prohibitive memory consumption for the input across workers. Unlike these methods, \MixGCN proposes a novel approach to integrate feature-level parallelism with node-level parallelism, achieving both scalability and practicality.

Furthermore, we note that feature-level parallelism has been widely adopted in GPU kernels~\cite{rahman2021fusedmm,zhang2022understanding,fu2022tlpgnn} for GCN acceleration, as it helps avoid branch divergence and enhance data locality~\cite{zhang2023survey}. However, these kernels focus solely on the \textit{neighbor aggregation} process and neglect the \textit{node update} process, making it non-trivial to extend them to end-to-end distributed GCN training.

\niparagraph{GCN Accelerators} As discussed in Section~\ref{sec:background_acc}, GCN acceleration has been extensively explored in the architecture community~\cite{zhang2023survey}. However, we observe that the design of scalable training architectures for GCNs remains underinvestigated. \MixGCN highlights two distinct challenges in this regard: accelerating S-SpMM and addressing the scalability limitations of fine-grain pipelines. To address these challenges, we propose a dedicated accelerator design and a node reordering technique, respectively, which collectively enable efficient and scalable GCN training.

\subsection{Limitations and Future Work}

\subsubsection{MoP with a Giant Adjacency Matrix}

One limitation of our proposed \MoP is that it requires duplicating the propagation matrix $\widehat{A}$ across all accelerators for \textit{neighbor aggregation}, which could potentially lead to memory bottlenecks. 
We clarify that $\widehat{A}$ is in general not the memory bottleneck. For example, training a 3-layer GCN with 128 hidden units for ogbn-papers100M \cite{hu2020open} requires 301GB for storing intermediate embeddings and output logits,  
but only needs 24GB for storing the adjacency matrix $\widehat{A}$, which is affordable to modern accelerators. 
Nevertheless, for extremely giant graphs where $\widehat{A}$ cannot fit into a single accelerator, alternative solutions are needed. 
One potential approach is to combine partition parallelism with \MoP, leveraging \MoP to accelerate intra-partition computations. 
Another direction for handling a giant $\widehat{A}$ is to distribute its storage across multiple accelerators where each accelerator maintains a portion of $\widehat{A}$ and fetch the remaining portions sequentially from other accelerators during its computation, thereby distributing the memory overhead of $\widehat{A}$. Exploring these designs and developing efficient methods for handling giant adjacency matrices remains an area for future research.

\niparagraph{No Support for Sophisticated GNN Models} Another limitation of our proposed \MoP is that it is constrained to handling element-wise reduction due to the inherent restrictions of feature-level parallelism. This limitation prevents \MoP from supporting more complex aggregation methods, such as those employed in graph attention networks~\cite{velivckovic2017graph}. However, it is worth noting that GCNs relying on element-wise aggregation remain the state-of-the-art for large-scale graph training tasks~\cite{hu2020open}, making \MoP a suitable solution for the majority of large-scale GCN workloads. Future research directions could involve extending \MoP to accommodate more sophisticated aggregation methods, enabling its application to a broader range of GNN architectures.

\niparagraph{Imbalanced Latency} As illustrated in Figure \ref{fig:breakdown}, we observe that the processing latency for \textit{neighbor aggregation} and \textit{node update} can be significantly imbalanced, which limits the achievable performance of fine-grain pipelines, as demonstrated in Table \ref{tab:exp_reorder}. One potential solution to mitigate this issue is to allocate different numbers of accelerators for each task, allowing for more efficient resource utilization. In our future work, we plan to explore dynamic resource allocation strategies and task-specific accelerator configurations.
\section{Conclusion}

Large-scale GCN training presents two significant challenges: managing \textit{giant feature tensors} and efficiently handling \textit{hybrid sparse-dense operations}. \MixGCN addresses these challenges by seamlessly integrating \MoP and \MoA, respectively. This novel approach achieves remarkable scalability through constant communication volume and feature memory usage, balanced workload distribution, and enhanced hardware efficiency. Both our theoretical analysis and empirical results validate the superior scalability of \MixGCN, positioning it as a promising solution for large-scale GCN training.

\bibliographystyle{plain}
\bibliography{reference}

\begin{thebibliography}{10}

\bibitem{abadi2016tensorflow}
Mart{\'\i}n Abadi, Ashish Agarwal, Paul Barham, Eugene Brevdo, Zhifeng Chen, Craig Citro, Greg~S Corrado, Andy Davis, Jeffrey Dean, Matthieu Devin, et~al.
\newblock Tensorflow: Large-scale machine learning on heterogeneous distributed systems.
\newblock {\em arXiv preprint arXiv:1603.04467}, 2016.

\bibitem{auten2020hardware}
Adam Auten, Matthew Tomei, and Rakesh Kumar.
\newblock Hardware acceleration of graph neural networks.
\newblock In {\em 2020 57th ACM/IEEE Design Automation Conference (DAC)}, pages 1--6. IEEE, 2020.

\bibitem{bai2023staleness}
Guangji Bai, Ziyang Yu, Zheng Chai, Yue Cheng, and Liang Zhao.
\newblock Staleness-alleviated distributed gnn training via online dynamic-embedding prediction.
\newblock {\em arXiv preprint arXiv:2308.13466}, 2023.

\bibitem{barham2022pathways}
Paul Barham, Aakanksha Chowdhery, Jeff Dean, Sanjay Ghemawat, Steven Hand, Daniel Hurt, Michael Isard, Hyeontaek Lim, Ruoming Pang, Sudip Roy, et~al.
\newblock Pathways: Asynchronous distributed dataflow for ml.
\newblock {\em Proceedings of Machine Learning and Systems}, 4:430--449, 2022.

\bibitem{chai2022distributed}
Zheng Chai, Guangji Bai, Liang Zhao, and Yue Cheng.
\newblock Distributed graph neural network training with periodic stale representation synchronization.
\newblock {\em arXiv preprint arXiv:2206.00057}, 2022.

\bibitem{chen2021dygnn}
Cen Chen, Kenli Li, Xiaofeng Zou, and Yangfan Li.
\newblock Dygnn: Algorithm and architecture support of dynamic pruning for graph neural networks.
\newblock In {\em 2021 58th ACM/IEEE Design Automation Conference (DAC)}, pages 1201--1206. IEEE, 2021.

\bibitem{chen2020graph}
Fenxiao Chen, Yun-Cheng Wang, Bin Wang, and C-C~Jay Kuo.
\newblock Graph representation learning: a survey.
\newblock {\em APSIPA Transactions on Signal and Information Processing}, 9, 2020.

\bibitem{chen2015mxnet}
Tianqi Chen, Mu~Li, Yutian Li, Min Lin, Naiyan Wang, Minjie Wang, Tianjun Xiao, Bing Xu, Chiyuan Zhang, and Zheng Zhang.
\newblock Mxnet: A flexible and efficient machine learning library for heterogeneous distributed systems.
\newblock {\em arXiv preprint arXiv:1512.01274}, 2015.

\bibitem{chen2021rubik}
Xiaobing Chen, Yuke Wang, Xinfeng Xie, Xing Hu, Abanti Basak, Ling Liang, Mingyu Yan, Lei Deng, Yufei Ding, Zidong Du, et~al.
\newblock Rubik: A hierarchical architecture for efficient graph neural network training.
\newblock {\em IEEE Transactions on Computer-Aided Design of Integrated Circuits and Systems}, 2021.

\bibitem{chinn1982bandwidth}
Phyllis~Z Chinn, Jarmila Chv{\'a}talov{\'a}, Alexander~K Dewdney, and Norman~E Gibbs.
\newblock The bandwidth problem for graphs and matrices—a survey.
\newblock {\em Journal of Graph Theory}, 6(3):223--254, 1982.

\bibitem{cuthill1969reducing}
Elizabeth Cuthill and James McKee.
\newblock Reducing the bandwidth of sparse symmetric matrices.
\newblock In {\em Proceedings of the 1969 24th national conference}, pages 157--172, 1969.

\bibitem{dave2021hardware}
Shail Dave, Riyadh Baghdadi, Tony Nowatzki, Sasikanth Avancha, Aviral Shrivastava, and Baoxin Li.
\newblock Hardware acceleration of sparse and irregular tensor computations of ml models: A survey and insights.
\newblock {\em Proceedings of the IEEE}, 109(10):1706--1752, 2021.

\bibitem{fedus2021switch}
William Fedus, Barret Zoph, and Noam Shazeer.
\newblock Switch transformers: Scaling to trillion parameter models with simple and efficient sparsity.
\newblock {\em arXiv preprint arXiv:2101.03961}, 2021.

\bibitem{fey2021gnnautoscale}
Matthias Fey, Jan~E Lenssen, Frank Weichert, and Jure Leskovec.
\newblock Gnnautoscale: Scalable and expressive graph neural networks via historical embeddings.
\newblock {\em arXiv preprint arXiv:2106.05609}, 2021.

\bibitem{fu2022tlpgnn}
Qiang Fu, Yuede Ji, and H~Howie Huang.
\newblock Tlpgnn: A lightweight two-level parallelism paradigm for graph neural network computation on gpu.
\newblock In {\em Proceedings of the 31st International Symposium on High-Performance Parallel and Distributed Computing}, pages 122--134, 2022.

\bibitem{gandhi2021p3}
Swapnil Gandhi and Anand~Padmanabha Iyer.
\newblock P3: Distributed deep graph learning at scale.
\newblock In {\em 15th {USENIX} Symposium on Operating Systems Design and Implementation ({OSDI} 21)}, pages 551--568, 2021.

\bibitem{garg2021understanding}
Raveesh Garg, Eric Qin, Francisco Mu{\~n}oz-Mart{\'\i}nez, Robert Guirado, Akshay Jain, Sergi Abadal, Jos{\'e}~L Abell{\'a}n, Manuel~E Acacio, Eduard Alarc{\'o}n, Sivasankaran Rajamanickam, et~al.
\newblock Understanding the design space of sparse/dense multiphase dataflows for mapping graph neural networks on spatial accelerators.
\newblock {\em arXiv preprint arXiv:2103.07977}, 2021.

\bibitem{geng2020awb}
Tong Geng, Ang Li, Runbin Shi, Chunshu Wu, Tianqi Wang, Yanfei Li, Pouya Haghi, Antonino Tumeo, Shuai Che, Steve Reinhardt, et~al.
\newblock Awb-gcn: A graph convolutional network accelerator with runtime workload rebalancing.
\newblock In {\em 2020 53rd Annual IEEE/ACM International Symposium on Microarchitecture (MICRO)}, pages 922--936. IEEE, 2020.

\bibitem{geng2021gcn}
Tong Geng, Chunshu Wu, Yongan Zhang, Cheng Tan, Chenhao Xie, Haoran You, Martin Herbordt, Yingyan Lin, and Ang Li.
\newblock I-gcn: A graph convolutional network accelerator with runtime locality enhancement through islandization.
\newblock In {\em MICRO-54: 54th Annual IEEE/ACM International Symposium on Microarchitecture}, pages 1051--1063, 2021.

\bibitem{guo2022dataefficient}
Minghao Guo, Veronika Thost, Beichen Li, Payel Das, Jie Chen, and Wojciech Matusik.
\newblock Data-efficient graph grammar learning for molecular generation.
\newblock In {\em International Conference on Learning Representations}, 2022.

\bibitem{hamilton2017inductive}
Will Hamilton, Zhitao Ying, and Jure Leskovec.
\newblock Inductive representation learning on large graphs.
\newblock In {\em Advances in neural information processing systems}, pages 1024--1034, 2017.

\bibitem{harlap2018pipedream}
Aaron Harlap, Deepak Narayanan, Amar Phanishayee, Vivek Seshadri, Nikhil Devanur, Greg Ganger, and Phil Gibbons.
\newblock Pipedream: Fast and efficient pipeline parallel dnn training.
\newblock {\em arXiv preprint arXiv:1806.03377}, 2018.

\bibitem{hu2020open}
Weihua Hu, Matthias Fey, Marinka Zitnik, Yuxiao Dong, Hongyu Ren, Bowen Liu, Michele Catasta, and Jure Leskovec.
\newblock Open graph benchmark: Datasets for machine learning on graphs.
\newblock {\em arXiv preprint arXiv:2005.00687}, 2020.

\bibitem{huang2020recurrent}
Jiyu Huang, Lin Guan, Yinsheng Su, Haicheng Yao, Mengxuan Guo, and Zhi Zhong.
\newblock Recurrent graph convolutional network-based multi-task transient stability assessment framework in power system.
\newblock {\em IEEE Access}, 8:93283--93296, 2020.

\bibitem{huang2024wisegraph}
Kezhao Huang, Jidong Zhai, Liyan Zheng, Haojie Wang, Yuyang Jin, Qihao Zhang, Runqing Zhang, Zhen Zheng, Youngmin Yi, and Xipeng Shen.
\newblock Wisegraph: Optimizing gnn with joint workload partition of graph and operations.
\newblock In {\em Proceedings of the Nineteenth European Conference on Computer Systems}, pages 1--17, 2024.

\bibitem{huang2019gpipe}
Yanping Huang, Youlong Cheng, Ankur Bapna, Orhan Firat, Dehao Chen, Mia Chen, HyoukJoong Lee, Jiquan Ngiam, Quoc~V Le, Yonghui Wu, et~al.
\newblock Gpipe: Efficient training of giant neural networks using pipeline parallelism.
\newblock In {\em Advances in neural information processing systems}, pages 103--112, 2019.

\bibitem{jeon2024graphpipe}
Byungsoo Jeon, Mengdi Wu, Shiyi Cao, Sunghyun Kim, Sunghyun Park, Neeraj Aggarwal, Colin Unger, Daiyaan Arfeen, Peiyuan Liao, Xupeng Miao, et~al.
\newblock Graphpipe: Improving performance and scalability of dnn training with graph pipeline parallelism.
\newblock {\em arXiv preprint arXiv:2406.17145}, 2024.

\bibitem{ji2021survey}
Shaoxiong Ji, Shirui Pan, Erik Cambria, Pekka Marttinen, and S~Yu Philip.
\newblock A survey on knowledge graphs: Representation, acquisition, and applications.
\newblock {\em IEEE Transactions on Neural Networks and Learning Systems}, 2021.

\bibitem{jia2020improving}
Zhihao Jia, Sina Lin, Mingyu Gao, Matei Zaharia, and Alex Aiken.
\newblock Improving the accuracy, scalability, and performance of graph neural networks with roc.
\newblock {\em Proceedings of Machine Learning and Systems (MLSys)}, pages 187--198, 2020.

\bibitem{jiang2013survey}
Chuntao Jiang, Frans Coenen, and Michele Zito.
\newblock A survey of frequent subgraph mining algorithms.
\newblock {\em The Knowledge Engineering Review}, 28(1):75--105, 2013.

\bibitem{jiang2020unified}
Yimin Jiang, Yibo Zhu, Chang Lan, Bairen Yi, Yong Cui, and Chuanxiong Guo.
\newblock A unified architecture for accelerating distributed $\{$DNN$\}$ training in heterogeneous $\{$GPU/CPU$\}$ clusters.
\newblock In {\em 14th USENIX Symposium on Operating Systems Design and Implementation (OSDI 20)}, pages 463--479, 2020.

\bibitem{jouppi2017datacenter}
Norman~P Jouppi, Cliff Young, Nishant Patil, David Patterson, Gaurav Agrawal, Raminder Bajwa, Sarah Bates, Suresh Bhatia, Nan Boden, Al~Borchers, et~al.
\newblock In-datacenter performance analysis of a tensor processing unit.
\newblock In {\em Proceedings of the 44th annual international symposium on computer architecture}, pages 1--12, 2017.

\bibitem{karypis1998fast}
George Karypis and Vipin Kumar.
\newblock A fast and high quality multilevel scheme for partitioning irregular graphs.
\newblock {\em SIAM Journal on scientific Computing}, 20(1):359--392, 1998.

\bibitem{kiningham2020grip}
Kevin Kiningham, Christopher Re, and Philip Levis.
\newblock Grip: A graph neural network accelerator architecture.
\newblock {\em arXiv preprint arXiv:2007.13828}, 2020.

\bibitem{kipf2016semi}
Thomas~N Kipf and Max Welling.
\newblock Semi-supervised classification with graph convolutional networks.
\newblock {\em arXiv preprint arXiv:1609.02907}, 2016.

\bibitem{kwak2010twitter}
Haewoon Kwak, Changhyun Lee, Hosung Park, and Sue Moon.
\newblock What is twitter, a social network or a news media?
\newblock In {\em Proceedings of the 19th international conference on World wide web}, pages 591--600, 2010.

\bibitem{kwon2018maeri}
Hyoukjun Kwon, Ananda Samajdar, and Tushar Krishna.
\newblock Maeri: Enabling flexible dataflow mapping over dnn accelerators via reconfigurable interconnects.
\newblock {\em ACM SIGPLAN Notices}, 53(2):461--475, 2018.

\bibitem{lepikhin2020gshard}
Dmitry Lepikhin, HyoukJoong Lee, Yuanzhong Xu, Dehao Chen, Orhan Firat, Yanping Huang, Maxim Krikun, Noam Shazeer, and Zhifeng Chen.
\newblock Gshard: Scaling giant models with conditional computation and automatic sharding.
\newblock {\em arXiv preprint arXiv:2006.16668}, 2020.

\bibitem{li2021gcnax}
Jiajun Li, Ahmed Louri, Avinash Karanth, and Razvan Bunescu.
\newblock Gcnax: A flexible and energy-efficient accelerator for graph convolutional neural networks.
\newblock In {\em 2021 IEEE International Symposium on High-Performance Computer Architecture (HPCA)}, pages 775--788. IEEE, 2021.

\bibitem{li2020pytorch}
Shen Li, Yanli Zhao, Rohan Varma, Omkar Salpekar, Pieter Noordhuis, Teng Li, Adam Paszke, Jeff Smith, Brian Vaughan, Pritam Damania, et~al.
\newblock Pytorch distributed: Experiences on accelerating data parallel training.
\newblock {\em arXiv preprint arXiv:2006.15704}, 2020.

\bibitem{li2021terapipe}
Zhuohan Li, Siyuan Zhuang, Shiyuan Guo, Danyang Zhuo, Hao Zhang, Dawn Song, and Ion Stoica.
\newblock Terapipe: Token-level pipeline parallelism for training large-scale language models.
\newblock In {\em International Conference on Machine Learning}, pages 6543--6552. PMLR, 2021.

\bibitem{liang2020engn}
Shengwen Liang, Ying Wang, Cheng Liu, Lei He, LI~Huawei, Dawen Xu, and Xiaowei Li.
\newblock Engn: A high-throughput and energy-efficient accelerator for large graph neural networks.
\newblock {\em IEEE Transactions on Computers}, 70(9):1511--1525, 2020.

\bibitem{lindholm2008nvidia}
Erik Lindholm, John Nickolls, Stuart Oberman, and John Montrym.
\newblock Nvidia tesla: A unified graphics and computing architecture.
\newblock {\em IEEE micro}, 28(2):39--55, 2008.

\bibitem{lu2017flexflow}
Wenyan Lu, Guihai Yan, Jiajun Li, Shijun Gong, Yinhe Han, and Xiaowei Li.
\newblock Flexflow: A flexible dataflow accelerator architecture for convolutional neural networks.
\newblock In {\em 2017 IEEE International Symposium on High Performance Computer Architecture (HPCA)}, pages 553--564. IEEE, 2017.

\bibitem{ma2019neugraph}
Lingxiao Ma, Zhi Yang, Youshan Miao, Jilong Xue, Ming Wu, Lidong Zhou, and Yafei Dai.
\newblock {NeuGraph}: Parallel deep neural network computation on large graphs.
\newblock In {\em 2019 USENIX Annual Technical Conference (USENIX ATC 19)}, pages 443--458, 2019.

\bibitem{massini2003all}
Annalisa Massini.
\newblock All-to-all personalized communication on multistage interconnection networks.
\newblock {\em Discrete applied mathematics}, 128(2-3):435--446, 2003.

\bibitem{md2021distgnn}
Vasimuddin Md, Sanchit Misra, Guixiang Ma, Ramanarayan Mohanty, Evangelos Georganas, Alexander Heinecke, Dhiraj Kalamkar, Nesreen~K Ahmed, and Sasikanth Avancha.
\newblock Distgnn: Scalable distributed training for large-scale graph neural networks.
\newblock In {\em Proceedings of the International Conference for High Performance Computing, Networking, Storage and Analysis}, pages 1--14, 2021.

\bibitem{mirhoseini2017device}
Azalia Mirhoseini, Hieu Pham, Quoc~V Le, Benoit Steiner, Rasmus Larsen, Yuefeng Zhou, Naveen Kumar, Mohammad Norouzi, Samy Bengio, and Jeff Dean.
\newblock Device placement optimization with reinforcement learning.
\newblock In {\em International Conference on Machine Learning}, pages 2430--2439. PMLR, 2017.

\bibitem{narayanan2019pipedream}
Deepak Narayanan, Aaron Harlap, Amar Phanishayee, Vivek Seshadri, Nikhil~R Devanur, Gregory~R Ganger, Phillip~B Gibbons, and Matei Zaharia.
\newblock Pipedream: generalized pipeline parallelism for dnn training.
\newblock In {\em Proceedings of the 27th ACM Symposium on Operating Systems Principles}, pages 1--15, 2019.

\bibitem{peng2022sancus}
Jingshu Peng, Zhao Chen, Yingxia Shao, Yanyan Shen, Lei Chen, and Jiannong Cao.
\newblock Sancus: staleness-aware communication-avoiding full-graph decentralized training in large-scale graph neural networks.
\newblock {\em Proceedings of the VLDB Endowment}, 15(9):1937--1950, 2022.

\bibitem{rahman2021fusedmm}
Md~Khaledur Rahman, Majedul~Haque Sujon, and Ariful Azad.
\newblock Fusedmm: A unified sddmm-spmm kernel for graph embedding and graph neural networks.
\newblock In {\em 2021 IEEE International Parallel and Distributed Processing Symposium (IPDPS)}, pages 256--266. IEEE, 2021.

\bibitem{rajbhandari2022deepspeed}
Samyam Rajbhandari, Conglong Li, Zhewei Yao, Minjia Zhang, Reza~Yazdani Aminabadi, Ammar~Ahmad Awan, Jeff Rasley, and Yuxiong He.
\newblock Deepspeed-moe: Advancing mixture-of-experts inference and training to power next-generation ai scale.
\newblock {\em arXiv preprint arXiv:2201.05596}, 2022.

\bibitem{rajbhandari2020zero}
Samyam Rajbhandari, Jeff Rasley, Olatunji Ruwase, and Yuxiong He.
\newblock Zero: Memory optimizations toward training trillion parameter models.
\newblock In {\em SC20: International Conference for High Performance Computing, Networking, Storage and Analysis}, pages 1--16. IEEE, 2020.

\bibitem{ramezani2021learn}
Morteza Ramezani, Weilin Cong, Mehrdad Mahdavi, Mahmut~T Kandemir, and Anand Sivasubramaniam.
\newblock Learn locally, correct globally: A distributed algorithm for training graph neural networks.
\newblock {\em arXiv preprint arXiv:2111.08202}, 2021.

\bibitem{rasley2020deepspeed}
Jeff Rasley, Samyam Rajbhandari, Olatunji Ruwase, and Yuxiong He.
\newblock Deepspeed: System optimizations enable training deep learning models with over 100 billion parameters.
\newblock In {\em Proceedings of the 26th ACM SIGKDD International Conference on Knowledge Discovery \& Data Mining}, pages 3505--3506, 2020.

\bibitem{sahni1976algorithms}
Sartaj~K Sahni.
\newblock Algorithms for scheduling independent tasks.
\newblock {\em Journal of the ACM (JACM)}, 23(1):116--127, 1976.

\bibitem{sergeev2018horovod}
Alexander Sergeev and Mike Del~Balso.
\newblock Horovod: fast and easy distributed deep learning in tensorflow.
\newblock {\em arXiv preprint arXiv:1802.05799}, 2018.

\bibitem{shazeer2018mesh}
Noam Shazeer, Youlong Cheng, Niki Parmar, Dustin Tran, Ashish Vaswani, Penporn Koanantakool, Peter Hawkins, HyoukJoong Lee, Mingsheng Hong, Cliff Young, et~al.
\newblock Mesh-tensorflow: Deep learning for supercomputers.
\newblock {\em Advances in neural information processing systems}, 31, 2018.

\bibitem{shoeybi2019megatron}
Mohammad Shoeybi, Mostofa Patwary, Raul Puri, Patrick LeGresley, Jared Casper, and Bryan Catanzaro.
\newblock Megatron-lm: Training multi-billion parameter language models using model parallelism.
\newblock {\em arXiv preprint arXiv:1909.08053}, 2019.

\bibitem{DC}
Synopsys.
\newblock Synopsys design compiler.
\newblock \url{https://www.synopsys.com/implementation-and-signoff/rtl-synthesis-test/dc-ultra.html}.
\newblock Accessed: 2022-02-17.

\bibitem{thorpe2021dorylus}
John Thorpe, Yifan Qiao, Jonathan Eyolfson, Shen Teng, Guanzhou Hu, Zhihao Jia, Jinliang Wei, Keval Vora, Ravi Netravali, Miryung Kim, et~al.
\newblock Dorylus: affordable, scalable, and accurate gnn training with distributed cpu servers and serverless threads.
\newblock In {\em 15th USENIX Symposium on Operating Systems Design and Implementation (OSDI 21)}, pages 495--514, 2021.

\bibitem{tripathy2020reducing}
Alok Tripathy, Katherine Yelick, and Aydin Buluc.
\newblock Reducing communication in graph neural network training.
\newblock {\em arXiv preprint arXiv:2005.03300}, 2020.

\bibitem{velivckovic2017graph}
Petar Veli{\v{c}}kovi{\'c}, Guillem Cucurull, Arantxa Casanova, Adriana Romero, Pietro Lio, and Yoshua Bengio.
\newblock Graph attention networks.
\newblock {\em arXiv preprint arXiv:1710.10903}, 2017.

\bibitem{wan2023adaptive}
Borui Wan, Juntao Zhao, and Chuan Wu.
\newblock Adaptive message quantization and parallelization for distributed full-graph gnn training.
\newblock {\em Proceedings of Machine Learning and Systems}, 5, 2023.

\bibitem{wan2022bns}
Cheng Wan, Youjie Li, Ang Li, Nam~Sung Kim, and Yingyan Lin.
\newblock {BNS-GCN}: Efficient full-graph training of graph convolutional networks with partition-parallelism and random boundary node sampling.
\newblock {\em Fifth Conference on Machine Learning and Systems}, 2022.

\bibitem{wan2022pipegcn}
Cheng Wan, Youjie Li, Cameron~R. Wolfe, Anastasios Kyrillidis, Nam~Sung Kim, and Yingyan Lin.
\newblock Pipe{GCN}: Efficient full-graph training of graph convolutional networks with pipelined feature communication.
\newblock In {\em International Conference on Learning Representations}, 2022.

\bibitem{wan2024towards}
Zishen Wan, Che-Kai Liu, Hanchen Yang, Chaojian Li, Haoran You, Yonggan Fu, Cheng Wan, Tushar Krishna, Yingyan Lin, and Arijit Raychowdhury.
\newblock Towards cognitive ai systems: a survey and prospective on neuro-symbolic ai.
\newblock {\em arXiv preprint arXiv:2401.01040}, 2024.

\bibitem{wang2021flexgraph}
Lei Wang, Qiang Yin, Chao Tian, Jianbang Yang, Rong Chen, Wenyuan Yu, Zihang Yao, and Jingren Zhou.
\newblock Flexgraph: a flexible and efficient distributed framework for gnn training.
\newblock In {\em Proceedings of the Sixteenth European Conference on Computer Systems}, pages 67--82, 2021.

\bibitem{wang2019supporting}
Minjie Wang, Chien-chin Huang, and Jinyang Li.
\newblock Supporting very large models using automatic dataflow graph partitioning.
\newblock In {\em Proceedings of the Fourteenth EuroSys Conference 2019}, pages 1--17, 2019.

\bibitem{wang2019dgl}
Minjie Wang, Da~Zheng, Zihao Ye, Quan Gan, Mufei Li, Xiang Song, Jinjing Zhou, Chao Ma, Lingfan Yu, Yu~Gai, Tianjun Xiao, Tong He, George Karypis, Jinyang Li, and Zheng Zhang.
\newblock Deep graph library: A graph-centric, highly-performant package for graph neural networks.
\newblock {\em arXiv preprint arXiv:1909.01315}, 2019.

\bibitem{wang2022neutronstar}
Qiange Wang, Yanfeng Zhang, Hao Wang, Chaoyi Chen, Xiaodong Zhang, and Ge~Yu.
\newblock Neutronstar: distributed gnn training with hybrid dependency management.
\newblock In {\em Proceedings of the 2022 International Conference on Management of Data}, pages 1301--1315, 2022.

\bibitem{wang2020gnnadvisor}
Yuke Wang, Boyuan Feng, Gushu Li, Shuangchen Li, Lei Deng, Yuan Xie, and Yufei Ding.
\newblock Gnnadvisor: An adaptive and efficient runtime system for gnn acceleration on gpus.
\newblock {\em arXiv preprint arXiv:2006.06608}, 2020.

\bibitem{wijesinghe2021new}
Asiri Wijesinghe and Qing Wang.
\newblock A new perspective on" how graph neural networks go beyond weisfeiler-lehman?".
\newblock In {\em International Conference on Learning Representations}, 2021.

\bibitem{wu2020comprehensive}
Zonghan Wu, Shirui Pan, Fengwen Chen, Guodong Long, Chengqi Zhang, and S~Yu Philip.
\newblock A comprehensive survey on graph neural networks.
\newblock {\em IEEE transactions on neural networks and learning systems}, 32(1):4--24, 2020.

\bibitem{xia2021graph}
Feng Xia, Ke~Sun, Shuo Yu, Abdul Aziz, Liangtian Wan, Shirui Pan, and Huan Liu.
\newblock Graph learning: A survey.
\newblock {\em IEEE Transactions on Artificial Intelligence}, 2(2):109--127, 2021.

\bibitem{xu2018powerful}
Keyulu Xu, Weihua Hu, Jure Leskovec, and Stefanie Jegelka.
\newblock How powerful are graph neural networks?
\newblock {\em arXiv preprint arXiv:1810.00826}, 2018.

\bibitem{xu2021gspmd}
Yuanzhong Xu, HyoukJoong Lee, Dehao Chen, Blake Hechtman, Yanping Huang, Rahul Joshi, Maxim Krikun, Dmitry Lepikhin, Andy Ly, Marcello Maggioni, et~al.
\newblock Gspmd: general and scalable parallelization for ml computation graphs.
\newblock {\em arXiv preprint arXiv:2105.04663}, 2021.

\bibitem{yan2020hygcn}
Mingyu Yan, Lei Deng, Xing Hu, Ling Liang, Yujing Feng, Xiaochun Ye, Zhimin Zhang, Dongrui Fan, and Yuan Xie.
\newblock Hygcn: A gcn accelerator with hybrid architecture.
\newblock In {\em 2020 IEEE International Symposium on High Performance Computer Architecture (HPCA)}, pages 15--29. IEEE, 2020.

\bibitem{yang2012defining}
Jaewon Yang and Jure Leskovec.
\newblock Defining and evaluating network communities based on ground-truth.
\newblock In {\em Proceedings of the ACM SIGKDD workshop on mining data semantics}, pages 1--8, 2012.

\bibitem{yang2000optimal}
Yuanyuan Yang and Jianchao Wang.
\newblock Optimal all-to-all personalized exchange in self-routable multistage networks.
\newblock {\em IEEE Transactions on Parallel and Distributed Systems}, 11(3):261--274, 2000.

\bibitem{ying2018graph}
Rex Ying, Ruining He, Kaifeng Chen, Pong Eksombatchai, William~L Hamilton, and Jure Leskovec.
\newblock Graph convolutional neural networks for web-scale recommender systems.
\newblock In {\em Proceedings of the 24th ACM SIGKDD International Conference on Knowledge Discovery \& Data Mining}, pages 974--983, 2018.

\bibitem{you2021gcod}
Haoran You, Tong Geng, Yongan Zhang, Ang Li, and Yingyan Lin.
\newblock Gcod: Graph convolutional network acceleration via dedicated algorithm and accelerator co-design.
\newblock {\em arXiv preprint arXiv:2112.11594}, 2021.

\bibitem{zeng2020graphact}
Hanqing Zeng and Viktor Prasanna.
\newblock Graphact: Accelerating gcn training on cpu-fpga heterogeneous platforms.
\newblock In {\em Proceedings of the 2020 ACM/SIGDA International Symposium on Field-Programmable Gate Arrays}, pages 255--265, 2020.

\bibitem{zhang2020hardware}
Bingyi Zhang, Hanqing Zeng, and Viktor Prasanna.
\newblock Hardware acceleration of large scale gcn inference.
\newblock In {\em 2020 IEEE 31st International Conference on Application-specific Systems, Architectures and Processors (ASAP)}, pages 61--68. IEEE, 2020.

\bibitem{zhang2020autosync}
Hao Zhang, Yuan Li, Zhijie Deng, Xiaodan Liang, Lawrence Carin, and Eric Xing.
\newblock Autosync: Learning to synchronize for data-parallel distributed deep learning.
\newblock {\em Advances in Neural Information Processing Systems}, 33:906--917, 2020.

\bibitem{zhang2022understanding}
Hengrui Zhang, Zhongming Yu, Guohao Dai, Guyue Huang, Yufei Ding, Yuan Xie, and Yu~Wang.
\newblock Understanding gnn computational graph: A coordinated computation, io, and memory perspective.
\newblock {\em Proceedings of Machine Learning and Systems}, 4:467--484, 2022.

\bibitem{zhang2024sylvie}
Meng Zhang, Qinghao Hu, Cheng Wan, Haozhao Wang, Peng Sun, Yonggang Wen, and Tianwei Zhang.
\newblock Sylvie: 3d-adaptive and universal system for large-scale graph neural network training.
\newblock In {\em 2024 IEEE 40th International Conference on Data Engineering (ICDE)}, pages 3823--3836. IEEE, 2024.

\bibitem{zhang2023survey}
Shichang Zhang, Atefeh Sohrabizadeh, Cheng Wan, Zijie Huang, Ziniu Hu, Yewen Wang, Jason Cong, Yizhou Sun, et~al.
\newblock A survey on graph neural network acceleration: Algorithms, systems, and customized hardware.
\newblock {\em arXiv preprint arXiv:2306.14052}, 2023.

\bibitem{zhang2021g}
Yongan Zhang, Haoran You, Yonggan Fu, Tong Geng, Ang Li, and Yingyan Lin.
\newblock G-cos: Gnn-accelerator co-search towards both better accuracy and efficiency.
\newblock In {\em 2021 IEEE/ACM International Conference On Computer Aided Design (ICCAD)}, pages 1--9. IEEE, 2021.

\bibitem{zhao2023pytorch}
Yanli Zhao, Andrew Gu, Rohan Varma, Liang Luo, Chien-Chin Huang, Min Xu, Less Wright, Hamid Shojanazeri, Myle Ott, Sam Shleifer, et~al.
\newblock Pytorch fsdp: experiences on scaling fully sharded data parallel.
\newblock {\em arXiv preprint arXiv:2304.11277}, 2023.

\bibitem{zheng2020distdgl}
Da~Zheng, Chao Ma, Minjie Wang, Jinjing Zhou, Qidong Su, Xiang Song, Quan Gan, Zheng Zhang, and George Karypis.
\newblock Distdgl: distributed graph neural network training for billion-scale graphs.
\newblock In {\em 2020 IEEE/ACM 10th Workshop on Irregular Applications: Architectures and Algorithms (IA3)}, pages 36--44. IEEE, 2020.

\bibitem{zheng2022alpa}
Lianmin Zheng, Zhuohan Li, Hao Zhang, Yonghao Zhuang, Zhifeng Chen, Yanping Huang, Yida Wang, Yuanzhong Xu, Danyang Zhuo, Joseph~E Gonzalez, et~al.
\newblock Alpa: Automating inter-and intra-operator parallelism for distributed deep learning.
\newblock {\em arXiv preprint arXiv:2201.12023}, 2022.

\bibitem{zhu2019aligraph}
Rong Zhu, Kun Zhao, Hongxia Yang, Wei Lin, Chang Zhou, Baole Ai, Yong Li, and Jingren Zhou.
\newblock Aligraph: A comprehensive graph neural network platform.
\newblock {\em arXiv preprint arXiv:1902.08730}, 2019.

\end{thebibliography}

\appendix

\newpage

\onecolumn

\section{Proof of Proposition~\ref{prop:pp_balance}}
\label{appx:proof}

We reduce a well-known NP-Hard problem identical-machines scheduling~\cite{sahni1976algorithms} to balancing workload of partition parallelism to conclude the proof.

\begin{proof}
Suppose that we have a total of $m$ workers. Partition parallelism distributes the computation of all nodes across $m$ workers. Denote the assigned subgraph of worker $i$ as $\mathcal{G}_i=(\mathcal{V}_i,\mathcal{E}_i)$, then the total floating point operations (FLOPs) for the $l$-th layer is:
$$\sum_{v\in\mathcal{V}_i}\left(deg(v)d^{(l)}+d^{(l)}d^{(l+1)}\right)$$
where $deg(v)$ represents the degree of node $v$ and $d^{(l)}$ denotes the dimension of the $l$-th layer's input. The first term $deg(v)d^{(l)}$ is the operations for aggregating the features of $v$'s neighbors, and we need extra $d^{(l)}d^{(l+1)}$ operations for updating its feature.

We now define $C_v\triangleq d^{(l)}(d^{(l+1)}+deg(v))$. For balancing the workload of partition parallelism, the partitioned node set $\{\mathcal{V}_1,\cdots,\mathcal{V}_m\}$ should minimize the following objective:
$$\max_{i\in[m]}\sum_{v\in\mathcal{V}_i}C_v$$
which is a standard identical-machines scheduling problem and well-known to be NP-Hard~\cite{sahni1976algorithms}. Hence, balancing the computation workload of partition parallelism is NP-Hard.
\end{proof}

\end{document}